\pdfoutput=1
\documentclass[journal,twoside,web]{packages/ieeecolor}
\usepackage{packages/jsen}

\usepackage{times}
\usepackage{epsfig}
\usepackage{graphicx}
\usepackage{amsmath}
\usepackage{amssymb}
\usepackage{amsfonts}
\usepackage{subfigure} 
\usepackage[ruled]{algorithm}
\usepackage{algpseudocode}
\usepackage{tikz}
\usepackage{ctable}
\usepackage[export]{adjustbox}
\usepackage{import}
\usepackage{pgf}
\usepackage[bookmarks=true]{hyperref}

\usepackage{ulem}
\usepackage{tcolorbox}
\usepackage{wrapfig}
\definecolor{abstractbg}{rgb}{0.89804,0.94510,0.83137}

\makeatletter
\let\NAT@parse\undefined
\makeatother
\usepackage[numbers]{natbib}

\usepackage{packages/sparo_acronyms}
\usepackage{packages/sparo_math}
\usepackage{packages/sparo_SIunits}
\usepackage{packages/sparo_misc}
\usepackage{multirow}

\usepackage{soul,color}
\usepackage{verbatim} %

\newcommand{\bl}[1]{{\textcolor{blue}{#1}}}
\newcommand{\rl}[1]{{\textcolor{red}{#1}}}

\usepackage[symbol]{footmisc}

\title{TRIDENT: Efficient Triple-Task Learning of Dehazing, Depth, and Uncertainty Estimation for Underwater 3-D Robot Visual Perception}

\author{Geonmo Yang${}^{1}$ and Younggun Cho${}^{1*}$
\thanks{${}^{1}$ G. Yang and Y. Cho are with the Dept. Electr. and Comput. Eng., Inha University, S. Korea \texttt{(e-mail: ygm7422@inha.edu, yg.cho@inha.ac.kr)} }
}%

\begin{document}

\IEEEtitleabstractindextext{%
\fcolorbox{abstractbg}{abstractbg}{%
\begin{minipage}
    {\textwidth}%
    \begin{wrapfigure}[22]{r}{3.8in}%
        \includegraphics[width=3.8in]{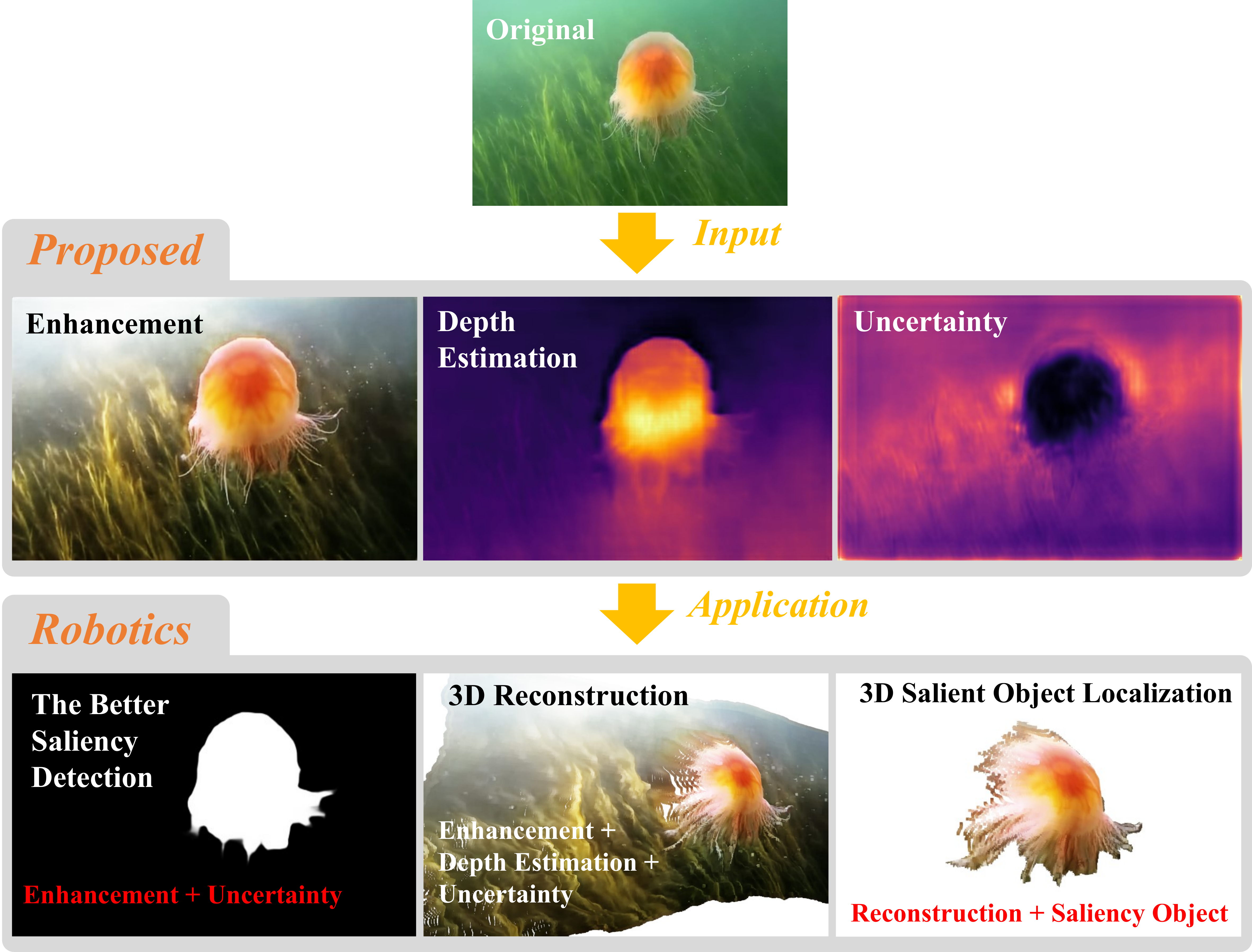}%
        \label{fig:thumbnail}
    \end{wrapfigure}%

\begin{abstract}
Underwater visual systems often suffer from blurry textures and low color contrast due to the inevitable light propagation. These issues can significantly degrade the perception of stable robotic operations. In this paper, we introduce a novel learning-based sensing system that tackles the multidimensional vision tasks in underwater; concretely, we deal with image enhancement, depth estimation, and uncertainty for 3-D visual systems. Also, we propose a \textbf{TRIDENT} model in a fast and lightweight manner; TRIDENT consists of three parallelized decoders and one backbone structure for efficient feature sharing. In addition, it is designed to be trained to express complex parameterization. In experimental evaluation on several standard datasets, we demonstrate that TRIDENT significantly outperforms other existing methods on image enhancement and depth estimation. Despite performing three tasks, our model has better efficiency than the others for both memory size and inference time. Finally, our joint learning approach demonstrates robustness in feature matching and seamlessly extends from 2-D to 3-D vision tasks. Supplements are available at \url{https://sites.google.com/view/underwater-trident/home}. 
\end{abstract}
\end{minipage}}
}

\maketitle

\section{Introduction}
\begin{figure*}[t!]
	\centering
    \def\width{1.0\textwidth}%
        {%
		\includegraphics[width=\width]{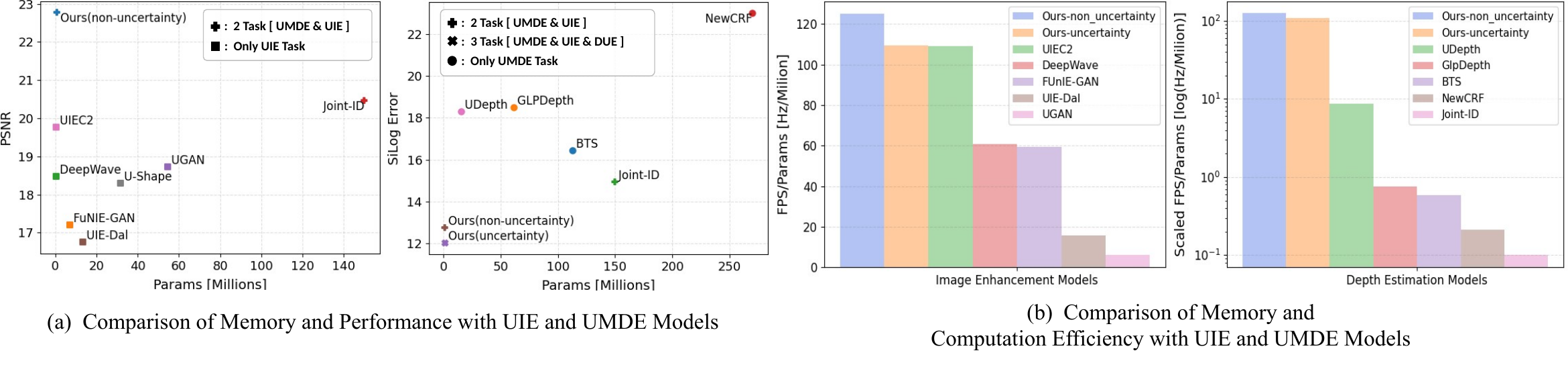}
		\label{fig:sample1_1}
	}
        \vspace{-6mm}
	\caption{Performance vs. model efficiency on Joint-ID Test set. All results are reported in the same and fair environments. (a) is a comparison graph of memory and performance with other UIE and UMDE models, and (b) is a comparison graph of memory and computation efficiency with other UIE and UMDE models. TRIDENT achieved state-of-the-art performance despite performing two or three tasks while being significantly more efficient than previous methods.}
	\label{fig:our_slam_figure}
	\vspace{-2mm}
\end{figure*}

As environmental issues such as marine debris and global warming increase in severity, the exploration and understanding of underwater environments through robotics have become imperative for various applications, including underwater scene analysis \cite{shkurti2012multi}, underwater mapping \cite{rahman2018sonar}, and 3-D object detection \cite{chen2021efficient}. 
However, the underwater environment poses significant challenges due to distortion and variable water depth, leading to color contrast loss and degradation of scene quality.

Notably, some studies \cite{darwiesh2020simulation, yang2020simulation} have highlighted the limitations of range sensors, such as \ac{LiDAR}, in underwater settings due to the inherent properties of water.
To ensure robust and accurate geometric operations in turbid underwater conditions, a multi-factor perception approach is imperative. This approach should encompass compact algorithm designs, address indistinct sensory data, and concurrently execute depth estimation and image enhancement processes. While existing literature addresses image enhancement \cite{sharma2023wavelength, peng2023u, islam2020fast, fabbri2018enhancing, uplavikar2019all, li2020underwater, wang2021uiec, islam2020simultaneous} and depth estimation \cite{yu2023udepth, yuan2022neural, kim2022global, lee2019big} as separate domains, it often lacks a unified approach. Even when an integrated system is offered, computational constraints for real-time applications are typically not considered. For example, several researchers \cite{islam2020fast, fabbri2018enhancing, yu2023udepth} have introduced systems with low computational demand for tasks such as restoration and depth estimation, leveraging learning-based approaches. However, these models often suffer from reduced perceptual accuracy or lower-resolution outputs. On the other hand, methodologies \cite{peng2023u, uplavikar2019all, wang2021uiec, yuan2022neural, kim2022global, lee2019big} that prioritize performance metrics tend to disregard computational and memory constraints, rendering them unsuitable for field robotics, where real-time processing is critical.
This dichotomy highlights the need for a holistic approach that harmoniously balances performance and computational efficiency in underwater perception research. In response to this gap, this paper introduces TRIDENT—a novel integrated network designed to perform three critical tasks simultaneously: \ac{UIE}, \ac{UMDE}, and \ac{DUE}. Motivated by the need for comprehensive solutions in the underwater environments, our contributions can be summarized as follows:

\begin{itemize}
    \item We present a novel approach and training pipeline with three interconnected objectives, enhancing geometric perception robustness.
    \item We propose TRIDENT, a lightweight and high-speed triple-task network strategically designed to address the specific challenges associated with each task (\figref{fig:our_slam_figure}).
    \item We prove TRIDENT outperforms recent \ac{SOTA} models in both image enhancement and depth estimation.
    \item We propose a novel pipeline to estimate perceptual instability while addressing performance degradation.
    \item From a robotics perspective, we showcase TRIDENT's utility across various robot vision tasks, emphasizing its versatility and applicability.
\end{itemize}

By jointly optimizing these tasks, our model enhances visibility and provides precise depth and depth uncertainty. This triple functionality extends the utility of our approach to 3-D vision tasks, enabling robotic systems to perceive and interact with their surroundings more comprehensively.

\section{Related Works}
\subsubsection{Underwater Image Enhancement} This is a critical area of research for reliable underwater vision processes. Model-based methods perform color correction using traditional image processing techniques or by estimating depth based on the amount of blurring. For example, \citet{galdran2015automatic} enhanced contrast by restoring short-wavelength colors, while \citet{berman2017diving} employed a haze line-based approach to estimate ambient color and transmission in color-biased environments. But, these models frequently face criticism for inadequate perceptual fidelity. In response, learning-based methods have significantly advanced the field. \citet{peng2023u} designed a transformer model called U-Shape with new network modules, employing a loss function across RGB, LAB, and LCH color spaces. Similarly, \citet{li2021underwater} proposed a learning pipeline that integrates features from different color spaces, such as HSV and LAB, in a network called UColor to enrich the diversity of feature representations. UIEC$^2$ \cite{wang2021uiec} also showed powerful performance by using both RGB and HSV space to consider the image hue. UIE-Dal \cite{uplavikar2019all} was designed with fully connected layers and a \ac{GAN} structure to achieve \ac{SOTA} performance. Despite these advances, these models primarily focus on natural color restoration without adequately addressing lightweight design and inference time. Some approaches, such as UGAN \cite{fabbri2018enhancing}, employ a \ac{GAN} architecture with a gradient difference loss function to enhance \ac{UIE} performance, while FUnIE-GAN \cite{islam2020fast} demonstrated fast inference times, making it suitable for embedded devices. However, these methods do not incorporate 3-D information.

\subsubsection{Monocular Depth Estimation} Dense depth maps are highly useful for geometric tasks such as navigation and 3-D vision. For this reason, monocular depth estimation has been extensively studied. Recently, \citet{zhang2023atlantis} attempted to train depth estimation models using diffusion to generate synthetic datasets, showing that these models generally perform better than traditional methodologies, but the improvement is marginal. In \cite{lee2019big}, \ac{LPG} was used to determine whether specific pixels are on the same plane, extending the process to multiple scales to capture both local and global features. \citet{yuan2022neural} used a transformer as an encoder and window Fully Connected-CRFs as a decoder to capture correlations between neighboring pixels. \citet{kim2022global} highlighted the robust capabilities of their Global-Local Path Networks Architecture that uses transformers and learns from local and global depth features. Recent studies \cite{yang2024depth, ke2024repurposing} have attempted to perform monocular depth estimation on large-scale datasets using diffusion or foundation models, achieving high generalization performance. However, most of them were studied in terrestrial environments and did not consider the degree of hazing and low color contrast in estimating depth. Furthermore, they did not consider model computation, which is important in robotics.

On the other hand, UDepth \cite{yu2023udepth} used pre-processed data to minimize the impact of water properties degradation, providing robust depth estimation. In addition, UDepth showed a fast inference time for robotics tasks. 

\subsubsection{Multi-Task Learning} There have been learning-based studies \cite{sharma2023wavelength, islam2020simultaneous, yang2023joint} that attempted to perform multiple tasks for robust computer vision tasks in underwater. DeepSESR \cite{islam2020simultaneous} proposed a robust learning-based vision framework that efficiently performs simultaneous image enhancement, super-resolution, and saliency detection. Similarly, DeepWave \cite{sharma2023wavelength} performed \ac{UIE} and super-resolution tasks simultaneously. 
Both methods did not consider depth information which is highly related to the level of image degradation.
Similar to our proposal, Joint-ID \cite{yang2023joint} performs depth estimation and image enhancement simultaneously. 
However, the model is unsuitable for real-time sensing due to the model size and only considers two tasks.

\section{Problem Formulation}
\subsection{Preliminary} \label{Preliminary}
Light attenuation and backscattering significantly influence underwater image degradation, typically modeled using McGlamery’s model \cite{wu2013simple} as follows:
\begin{equation} 
    I_{c}(x) = J_{c}(x) \cdot T_{c}(x) + B_{c} \cdot (1-T_{c}(x))
    \label{eq:distortion_model}
\end{equation}
, where $c \in \{ r, g, b \}$ is the wavelength component, $I_c(x)$ and $J_c(x)$ represent the observed and original images, respectively. $B_c$ is the homogeneous ambient light, and $T_c(x)$ is the medium transmission ratio. 
For a homogeneous medium, $T_c(x)$ is definable as follows:
\begin{equation} 
    T_c(x) = e^{-\beta_c \cdot d(x)}
    \label{eq:distortion_model_2}
\end{equation}
Here, $\beta_c$ is the attenuation coefficient with respect to the ambient medium, which is an invariant constant value, and $d(x)$ represents scene depth. $T_c(x)$ denotes the degree to which color intensity is transferred from the clean image. Consequently, Equation \eqref{eq:distortion_model} indicates that a complete image restoration is feasible only if we simultaneously obtain values for $T_c(x)$ and ambient light $B_c$. The restoration process using the inverse distortion model is defined as follows:
\begin{equation} 
    J_{c}(x) = \frac{I_{c}(x)}{T_{c}(x)} \; - \; \frac{B_c(1-T_{c}(x))}{T_{c}(x)}
    \label{eq:distortion_model_3}
\end{equation}
A detailed description of the restoration process based on this theory is provided in \ref{Enhancement Extraction Block}.

\vspace{-2mm}
\subsection{Training Pipeline} \label{section training overview}
The approach focuses on generating enhanced images from distorted inputs and estimating distance from the camera origin.
We formulate the problem as learning a pixel-to-pixel mapping from a source domain $X$ (of distorted images) to target domains $E$ (of enhanced images), D (of depth maps), and U (of depth uncertainty), representing this mapping as a generating function $G:X \rightarrow \{E, D, U\}$. To address inaccurate depth estimation and edge instability, we also estimate the depth uncertainty.

\section{Proposed Method}
\subsection{TRIDENT Network}
\begin{figure}[h!]
	\centering
	\def\width{1\columnwidth}%
    {%
		\includegraphics[width=\width]{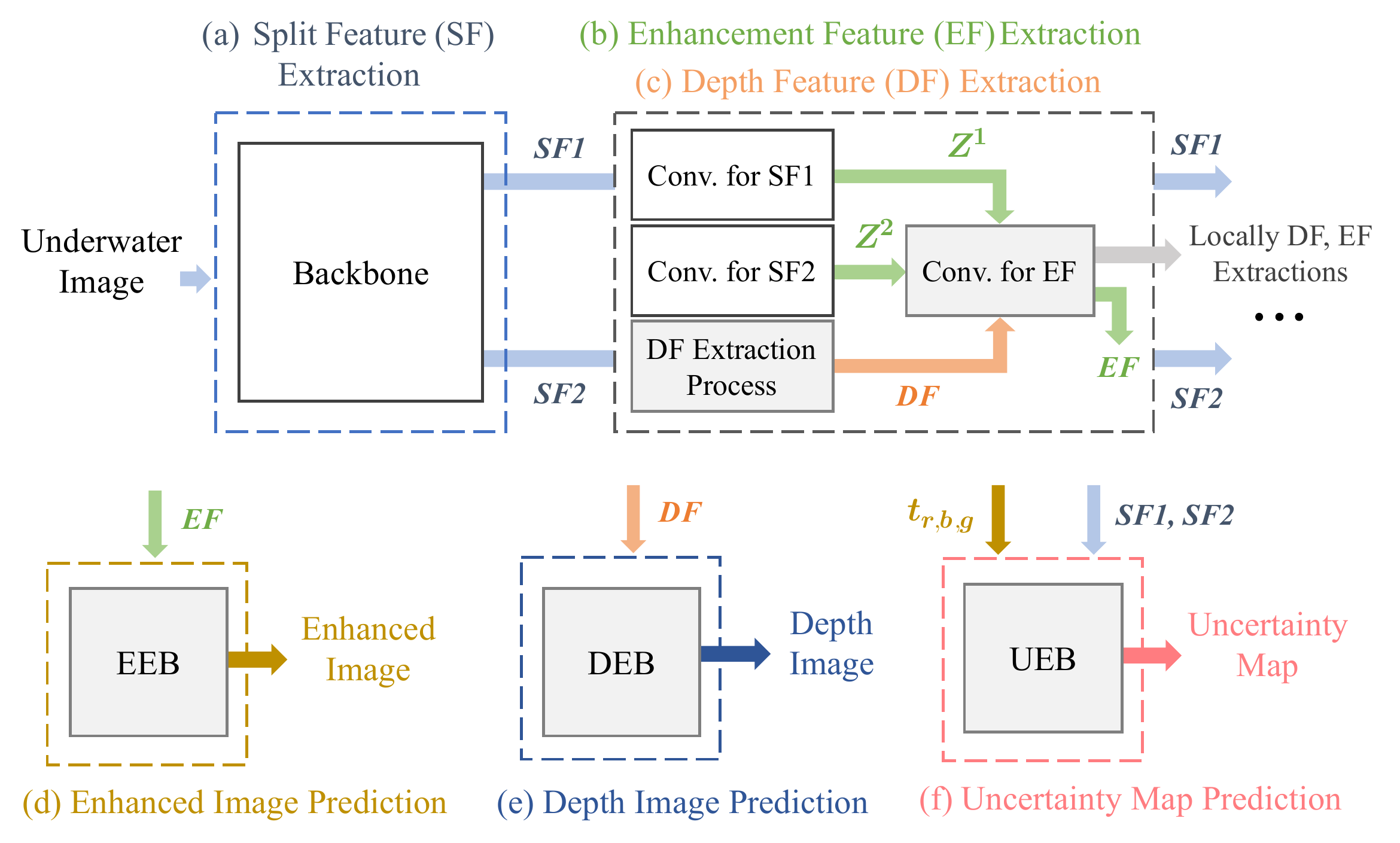}
	}
        \vspace{-5mm}
	\caption{The end-to-end architecture for TRIDENT process. The process is processed in six steps: (a) Split feature (SF) extraction, (b) enhancement feature (EF) extraction, (c) depth feature (DF) extraction, (d) enhanced image prediction, (e) depth map prediction, (f)} uncertainty estimation.
        \label{fig:overview}
\vspace{-3mm}
\end{figure}

\begin{table}[t!]
\caption{Abbreviations for network modules and pipeline courses table.}
\label{tab:abbreviations}
\centering
\resizebox{0.425\textwidth}{!}{%
\begin{tabular}{c||c|c}
\toprule
Type                                                                         & Full name                      & Abbreviations \\ \hline \hline
\multirow{7}{*}{\begin{tabular}[c]{@{}c@{}}Network   \\ Module\end{tabular}} & ConvBlock                      & CB            \\ \cline{2-3} 
                                                                             & UpConvBlock                    & UCB           \\ \cline{2-3} 
                                                                             & ConcatConvBlock                & CCB           \\ \cline{2-3} 
                                                                             & Depth Feature Extractor        & DFE           \\ \cline{2-3} 
                                                                             & Enhancement Extraction Block   & EEB           \\ \cline{2-3} 
                                                                             & Depth Extraction Block         & DEB           \\ \cline{2-3} 
                                                                             & Uncertainty Extraction Block   & UEB           \\ \hline
\multirow{3}{*}{Process}                                                     & Split Feature Extraction       & SF Extraction \\ \cline{2-3} 
                                                                             & Enhancement Feature Extraction & EF Extraction \\ \cline{2-3} 
                                                                             & Depth Feature Extraction       & DF Extraction \\
\bottomrule
\end{tabular}
}
\vspace{-4mm}
\end{table}

\begin{figure*}[t!]
	\centering
	\def\width{1.0\textwidth}%
        \includegraphics[width=\width]{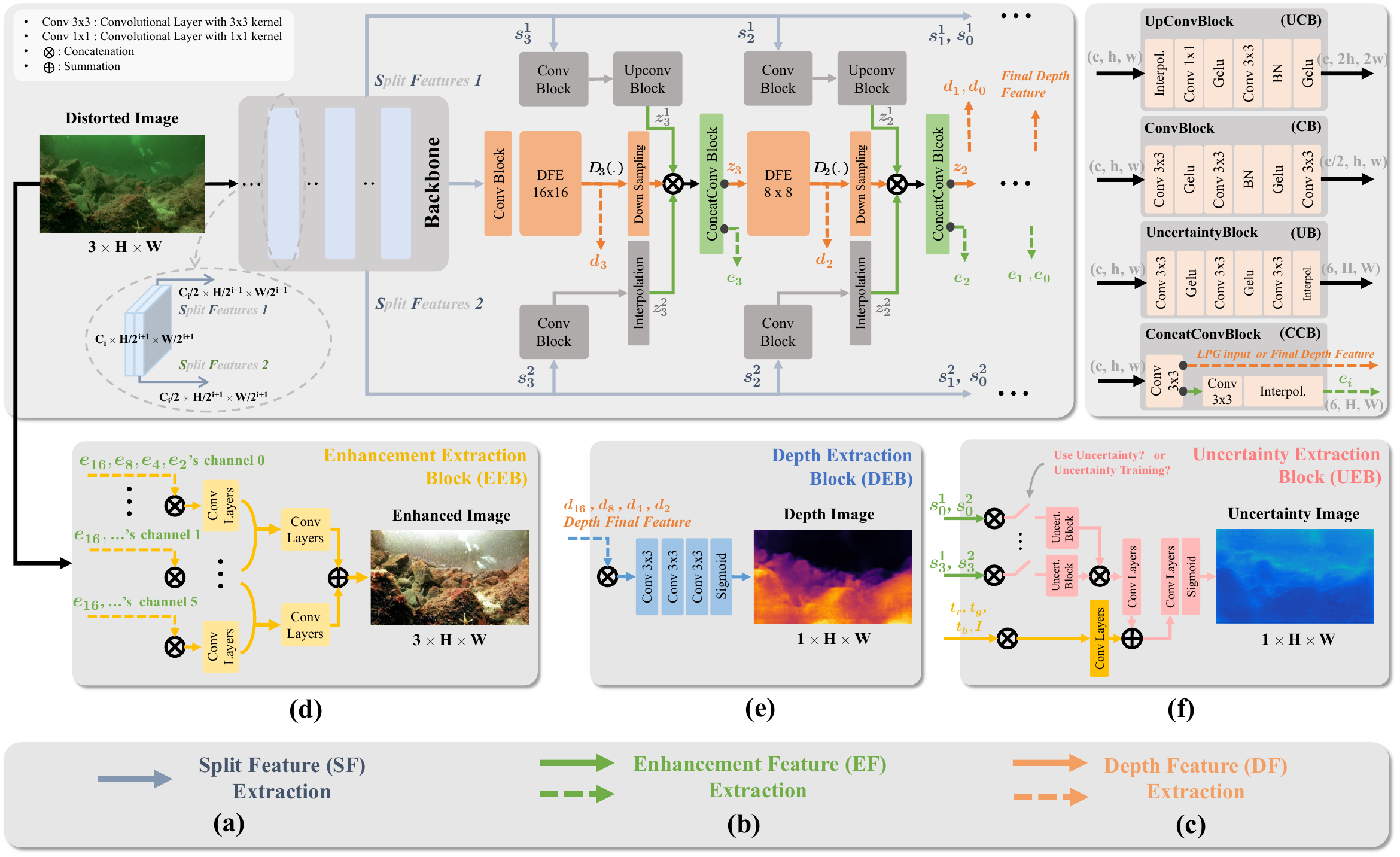}
        \vspace{-1mm}
	\caption{The end-to-end architecture for TRIDENT training. It shows the process of sharing hierarchical features based on the backbone and three-stage decoder. This figure shows an illustration of the components in TRIDENT, different block modules (UCB, CB, UB, and CCB), and the extraction blocks (EEB, DEB, and UEB) that take the objectives through each scale feature. (a) is the process of efficiently extracting features from the backbone. (b) and (c) are the processes of extracting enhancement and depth features. (d) and (e) are the processes of extracting the enhanced image and depth image through the features extracted in (b) and (c). Finally, (f) is the process of estimating uncertainty through the features extracted in (a) and the distortion information extracted in (d).}
        \label{fig:network}
\end{figure*}

The TRIDENT network is designed to address the challenges of robust depth estimation, image restoration, and perceptual uncertainty estimation in underwater environments. The architecture includes the backbone \cite{howard2019searching} and main components, as summarized in Table \ref{tab:abbreviations}.
The overall and detailed processes can be comprehensively understood through the six sequential executions in \figref{fig:overview} and \figref{fig:network}: SF Extraction, DF Extraction, EF Extraction, Depth Map Prediction, Enhanced Image Prediction, and Uncertainty Estimation.
\subsubsection{Split Feature (SF) Extraction - (a)} \label{sec:sf_extracion}

In this initial phase of the TRIDENT network, complex features are extracted from $k$-th layers in backbone \cite{howard2019searching}, where $k \in \{1, 3, 6, 12 \}$. The outputs, denoted as Split Feature (SF) $s_i \in \mathbb{R}^{C_i\times \frac{H}{2^{i+1}} \times \frac{W}{2^{i+1}}}$, where $i \in \{0, 1, 2, 3\}$ is feature index, are divided into two parallel branches features $s^1_i$ and $s^2_i \in \mathbb{R}^{\frac{C_i}{2} \times \frac{H}{2^{i+1}} \times \frac{W}{2^{i+1}}}$ to facilitate effective representation of the complex feature space in a lightweight manner. Here, $C_i$ represents the channels, and $H$, $W$ denotes the height and width of the distorted input image. These multi-scale compounded features are incorporated in the three-stage process for effective learning.
\subsubsection{Depth Feature (DF) Extraction - (c)} \label{sec:df-extraction} 
\begin{figure}[h!]
	\centering
	\def\width{1\columnwidth}%
        {%
		\includegraphics[width=\width]{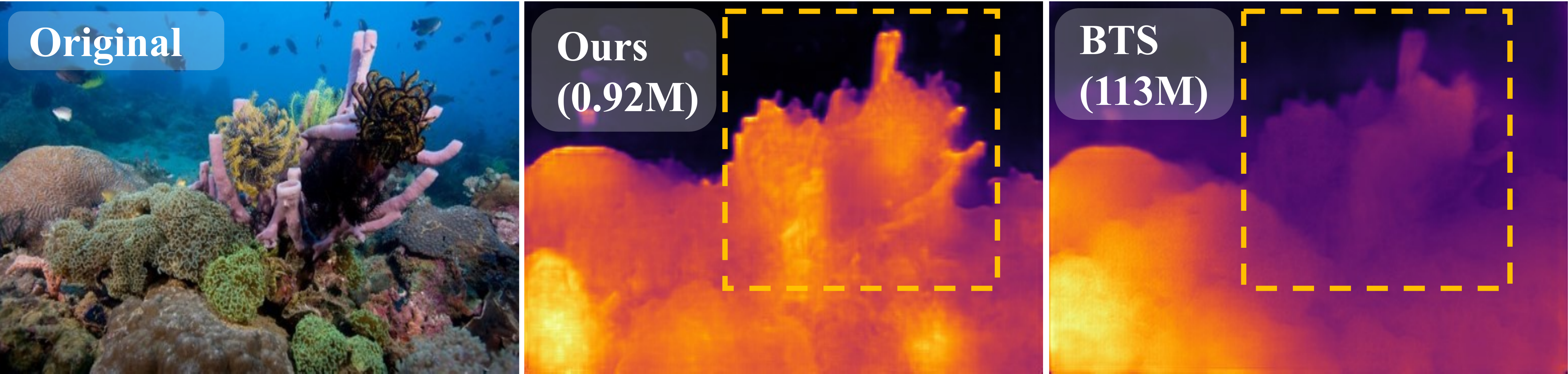}
	}
        \vspace{-4mm}
	\caption{The color maps of depth from light-weighted DFE in our network and original DFE in BTS. It shows that the model, which uses relatively light-weight modules, estimates depth more accurately.}
	\label{fig:depth_scale}
\end{figure}

The DF extraction process focuses on generating multi-scaled depth features (DF) $d_{i} \in \mathbb{R}^{1\times H \times W}$, while accounting for underwater distortion and hazing. Specifically, \figref{fig:network} shows an illustration of the CB, and DF extraction process is represented by the orange-colored blocks and arrows. 
Other methodologies \cite{lee2019big, yuan2022neural, kim2022global}, including those using the original DFE, often overlook the relevance of degradation and hazing and use heavy decoders for robust depth prediction. However, \citet{yang2023joint} demonstrates that depth feature extraction combining color degradation features in low-visibility environments can efficiently achieve accurate and robust depth feature extraction. Inspired by it, we use a modified DFE that drastically reduces the number of traditional contrast convolution layers while incorporating hazing features. Using this, we efficiently extracted normal vectors guided depth at each scale. \figref{fig:depth_scale} shows that result from this modified DFE is better than the original module result. Specifically, we employ CCB $B^{ccb}_{i}:= (\mathbb{R}^{\frac{C_{i}}{2}}, \mathbb{R}^{\frac{C_{i}}{2}}, \mathbb{R}^{1}) \mapsto (\mathbb{R}^{\frac{C_{i-1}}{2}},\mathbb{R}^6)$ so that depth features $d_i$ from DFE function $D_i := \mathbb{R}^{\frac{C_{i}}{2}} \mapsto \mathbb{R}^1$ can be mixed with feature $z^1_i$, $z^2_i$, which have complex attenuation and hazing features. This process can be expressed as follows:
\begin{equation}
    e_i, z_i = B^{ccb}_i(z^1_i, z^2_i, d_i), \;\;\; where \;\;\; d_i = D_i(z_{i+1}).
    \label{eq:feat_fusion_eq}
\end{equation} 
Here, $z_i$ represents the complex depth feature for next DEF function $D_{i-1}$.
The details of this process and $e_i, z^1_i, z^2_i$ are explained in \ref{sec:ef-extraction}. The final depth feature is the last embedding vector for collectively contributing to the creation of a competitive depth map.
\subsubsection{Enhancement Feature (EF) Extraction - (b)} \label{sec:ef-extraction} 
\begin{figure}[h!]
	\centering
	\def\width{1\columnwidth}%
        {%
		\includegraphics[width=\width]{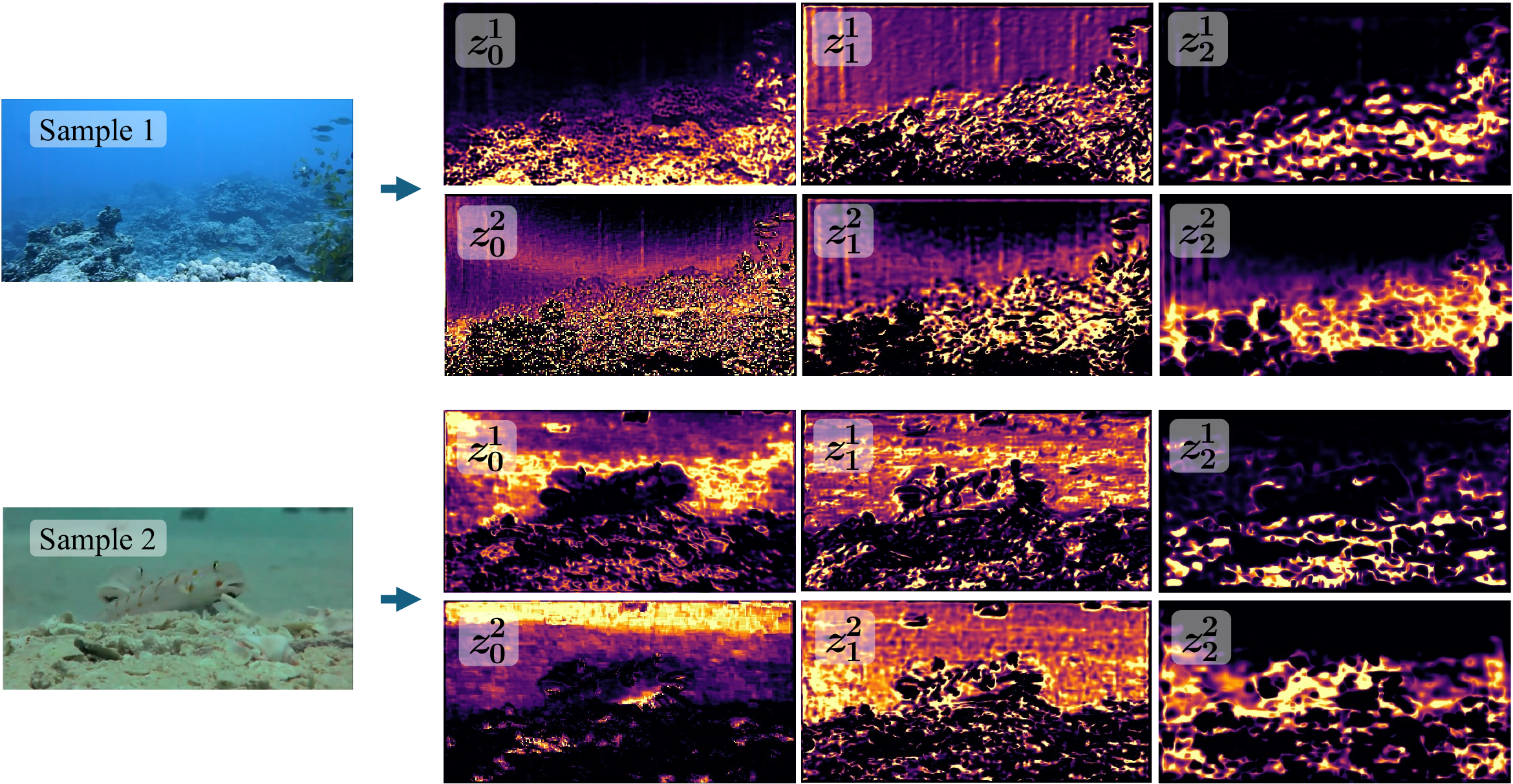}
	}
        \vspace{-4mm}
	\caption{Visualization samples of features $z^1_i$ and $z^2_i$. These figures, visualized as averages, illustrate either a proportional or inverse relationship depending on the distance.}
	\label{fig:feat_visual}
\end{figure}

EF Extraction involves the simultaneous extraction of Enhancement Feature (EF) $e_i \in \mathbb{R}^{6 \times H \times W}$ and $z_i$ from $s^1_i$, $s^2_i$, and $d_i$, which are input features of a function $D_{i-1}$. We accomplish it efficiently with a three-stage learning and integration strategy: first-stage learning uses each scaled feature $s^1_i$ and goes through CB function $B^{cb}_{i}:= \mathbb{R}^{\frac{C_{i}}{2}} \mapsto \mathbb{R}^{\frac{C_{i}}{2}}$ and UCB function $B^{ucb}_{i}:= \mathbb{R}^{\frac{C_{i}}{2}} \mapsto \mathbb{R}^{\frac{C_{i}}{2}}$, which are CNN layers. We define $z^1_i$ as the output of this process. Similarly, the second-stage learning takes $s^2_i$ as input and goes through CB function $B^{cb}_{i}$ and \texttt{BI} to generate compounded output $z^{2}_i$. The parallel CB and UCB are utilized to learn corresponding features. The last stage of learning is simply the DF extraction process. Unlike other models \cite{peng2023u, islam2020fast, fabbri2018enhancing} based on the U-net structure, it is designed by the hierarchical procedures to take the efficient feature-sharing process and light-weight manner. This structure allows our model to have low parameters while maintaining good performance. Finally, as mentioned \ref{sec:ef-extraction}, by utilizing the CCB function, we perform feature sharing for jointly learning from $z^1_i$, $z^{2}_i$, and $d_i$, composite features for depth and attenuation. The overall sharing process of CCB function $B^{ccb}_i$ can be represented by \figref{fig:network} and the following formula:
\begin{equation}
    e^{lr}_i, z_i = \texttt{Conv3x3}(z^1_i \otimes z^2_i \otimes d_i)
    \label{eq:conv1_eq}
\end{equation} 
\begin{equation}
    e_i = \texttt{BI}(\texttt{Conv3x3}(e^{lr}_i)).
    \label{eq:conv2_eq}
\end{equation}
According to equation \ref{eq:distortion_model_2}, the transmission value can be rewritten as:
\begin{equation}
    \log{T_c(x)} = -\beta_c \cdot d(x).
\end{equation} 
\begin{equation}
    \log \frac{1}{T_c(x)} \propto d(x).
\end{equation}
This means that the degree of hazing has a positive relationship with depth. \figref{fig:feat_visual} shows how $z^1_i$, $z^2_i$ generally proportional or inversely proportional to distance. This multi-feature fusion learning is more effective in understanding hazed scenes than physics-based learning methodologies \cite{cong2023pugan, chandrasekar2024phish} that do not do feature sharing.

\subsubsection{Enhancement Extraction Block (EEB) - (d)} \label{Enhancement Extraction Block} 
\begin{figure}[t!]
\vspace{2mm}
	\centering
	\def\width{1\columnwidth}%
        {%
		\includegraphics[width=\width]{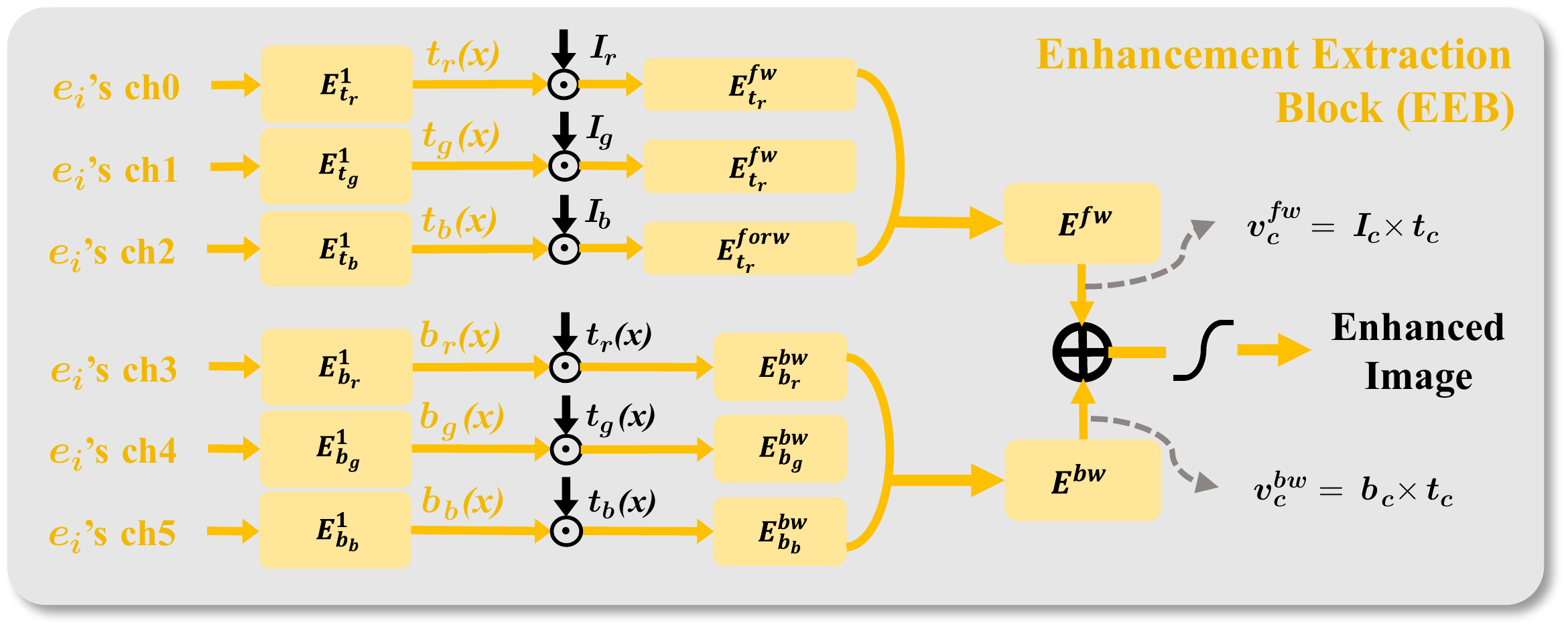}
	}
        \vspace{-6mm}
	\caption{Illustration of the EEB. It shows the detailed process in EEB for image restoration.}
	\label{fig:eeb_block}
	\vspace{-4mm}
\end{figure}

\begin{figure}[h!]
	\centering
	\def\width{1\columnwidth}%
        {%
		\includegraphics[width=\width]{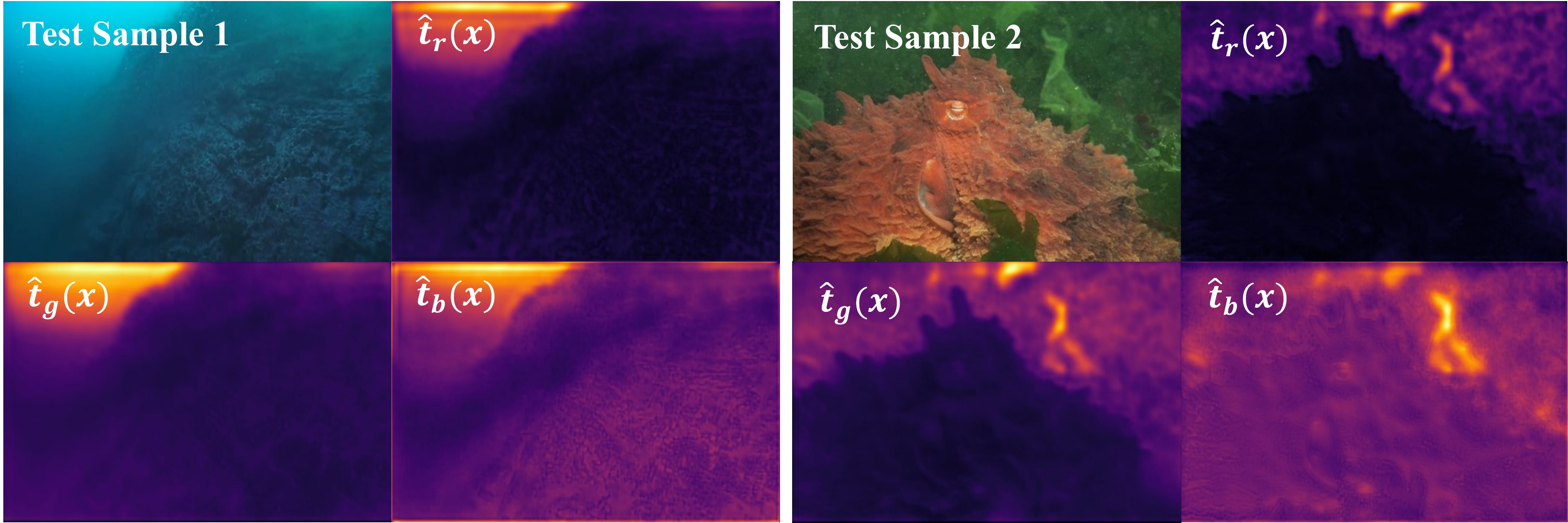}
	}
        \vspace{-4mm}
	\caption{The results of inverse transmission vector. It shows that the vectors have the property that the degree of distortion correction is increased with distance.}
	\label{fig:distortion_visualize}
\end{figure}

This is a restoration process with parameter backward inference. At first, we define two variables: variable $t_c(x)$ for forward-scattering restoration and variable $b_c(x)$ for back-scattering restoration. They can be expressed by Equation \eqref{eq:distortion_model_3} as
\begin{equation}
\begin{split}
    J_{c}(x) &= \frac{I_{c}(x)}{T_{c}(x)} \; - \; \frac{B_c(1-T_{c}(x))}{T_{c}(x)} \\
             &= \underbrace{I_{c}(x) \cdot t_{c}(x)}_{v^{fw}_c(x)}\;\; + \;\;\underbrace{b_c(x) \cdot t_{c}(x)}_{v^{bw}_c(x)}
    \label{eq:eeb_eq1}
\end{split}
\end{equation} 
, where $b_c(x) = B_c(T_{c}(x)-1)$ and $t_c(x) = \frac{1}{T_{c}(x)}$.
EEB is dedicated to generating perceptually enhanced images by estimating the embedding vectors $t_{c}$ and $b_c$ to obtain $v^{fw}_c \in \mathbb{R}^{3 \times H \times W}$ and $v^{bw}_c \in \mathbb{R}^{3\times H \times W}$. To consider the distortion model formula \cite{wu2013simple}, the enhanced image is extracted through a split-summation convolution operation. Specifically, the EEB consists of function $E^1_{t_c}:=\mathbb{R}^{4}\mapsto\mathbb{R}^{1}$, which estimates the inverse transmission embedding vector $t_c$ through multi-scaled features, function $E^1_{b_c}:=\mathbb{R}^{4}\mapsto\mathbb{R}^{1}$ CNN layer, which estimates the background distortion embedding vector $b_c$, and function $E^{fw}$, $E^{bw}$. After that, it goes through convolution layers in parallel. The detail of this process can be represented as follows and \figref{fig:eeb_block}: 
\begin{flalign}
& \quad \text{$forward$}
\;\;\;\left\{\begin{aligned}
    \;\;\; &\text{$t_\triangle = E^{1}_{t_\triangle}(e_{0,j}, e_{1,j}, e_{2,j}, e_{3,j})$} \\
    \;\;\; &\text{$v_c^{fw} = E^{fw}(v_r^{fw}, v_g^{fw}, v_b^{fw})$}
\end{aligned}
\right. &
\end{flalign}  
\begin{flalign}
& \quad \text{$backward$}
\;\;\left\{\begin{aligned}
    \;\;\; &\text{$b_\triangle = E^{1}_{b_\triangle}(e_{0,k}, e_{1,k}, e_{2,k}, e_{3,k})$} \\
    \;\;\; &\text{$v_c^{bw} = E^{bw}(v_r^{bw}, v_g^{bw}, v_b^{bw})$}
\end{aligned}
\right. &
\end{flalign}
, $j \in \{0, 1, 2\}$ and $k \in \{3, 4, 5\}$ are channel indexes. $v_\triangle^{fw}$ and $v_\triangle^{bw}$ are outputs from $E_{t_\triangle}(I_\triangle \cdot t_\triangle)$ and $E_{b_\triangle}(b_\triangle \cdot t_\triangle)$, respectively, where $\triangle \in \{r, g, b\}$ is the wavelength component. $E^{fw}$ and $E^{bw}$ are CNN layers for stable learning on \texttt{sigmoid}. Finally, the enhanced image is obtained by vector summation by equation (5). \figref{fig:distortion_visualize} shows the color maps for embedding vector $t_c$. It demonstrates that the vectors represent the degree of each hazing.
The transmission vector $t_c$, which represents both depth and attenuation, is used to estimate accurate uncertainty. 

\subsubsection{Depth Extraction Block (DEB) - (e)} DEB focuses on generating a comprehensive depth map from the feature $d_i$ inputs. This involves concatenating depth features at different scales and applying convolutional operations and \texttt{sigmoid} activation, ultimately producing a normalized depth image. The normalized depth image is multiplied by the maximum depth value to get the final depth map. The multi-scaled DF features are shown locally and globally in \figref{fig:depth_scale}. 
\subsubsection{Uncertainty Extraction Block (UEB) - (f)} \label{uncertainty extraction block}
Aleatoric uncertainty is widely used to assess data reliability, as demonstrated by \citet{kendall2017uncertainties}. However, previous studies \cite{liu2022modeling, valdenegro2022deeper} mention that it often leads unstable training process. In the context of our model, which operates within limited parameters and is applied to multi-task learning, this can result in instability and reduced performance under full capacity. To address this, we propose a novel uncertainty estimation method designed to mitigate these challenges and maintain model robustness. Information about depth and hazing can be used implicitly to estimate the degree of difficulty in recognition. UEB is introduced for 3-task training or scenarios requiring uncertainty analysis. The process uses the inverse transmission embedding vector and encoder features to estimate the region of the scene that is difficult to approximate. Utilizing $s^1_i$, $s^2_i$, and vector $t_c(x)$ as inputs, which have information about depth and hazing, the four-stage UB and CNN layers employ the three convolutional modules to extract uncertainty. 

\vspace{-3mm}
\subsection{Loss Function Formulation}
The end-to-end training of TRIDENT is supervised by two (not use uncertainty) or three (use uncertainty) loss components that address various aspects of learning function $G:X \rightarrow \{E, D, U \}$ (if not training UEB, exclude $U$). By denoting $\hat{E}, \hat{D}, \hat{U} = G(x)$ as the generated outputs, we formulate the loss term as follows.
\subsubsection{Loss Function for \ac{UIE}} For image enhancement, we used a supervised loss function called SSIM Loss \cite{zhao2016loss} to impose the structure and texture similarity on the enhanced image. It is defined as follows:
\begin{align} 
    &SSIM = \frac{2\mu_{\hat{I}}(\hat{x})\mu_{I}(x)+c_1}{\mu^2_{I}(x)+\mu^2_{\hat{I}}(\hat{x})+c_1}\cdot \frac{2\sigma_{I\hat{I}}(x,\hat{x})+c_2}{\sigma^2_I(x)+\sigma^2_{\hat{I}}(\hat{x})+c_2} \\[1pt]
    &\mathcal{L}_{SSIM}=1-\frac{1}{n}\sum_i SSIM(x_i,\hat{x}_i)
    \label{eq:ssim_loss}
\end{align}
, where $\mu_{\hat{I}}(\hat{x})$ and $\sigma_{\hat{I}}(\hat{x})$ correspond to the mean and standard deviation of the patch from the predicted image, similarly, $\mu_I(x)$ and $\sigma_I(x)$ are the same corresponds to ground truth image. And $\sigma_{I\hat{I}}(x,\hat{x})$ is cross-covariance. $n$ is the total number of pixels in the original image. We set $c_1$ = 0.02 and $c_2$ = 0.03. Finally, we derive $\mathcal{L}_{SSIM}$ by averaging the $SSIM$ for image enhancement. 
\subsubsection{Loss Function for \ac{UMDE}} For depth estimation, we use a popular Scale-Invariant Log (SILog) loss, introduced in \cite{eigen2014depth}. This formula is defined as follows:
\begin{equation} 
    \mathcal{L}_{SILog} = \sqrt{\frac{1}{n}\sum_i g_i^2-\lambda_{silog}(\sum_i g_i)^2}
    \label{eq:silog_loss}
\end{equation}
, where $g_i = log \hat{d}_i - log d_i$, and $\hat{d}_i$is the predicted depth, $d_i$is the ground truth depth. As in \citet{lee2019big}, we used $\lambda_{silog} = 0.85$ in our implementation.
\vspace{2mm}
\subsubsection{Loss Function for Depth Uncertainty} Uncertainty is implicitly derived from the depth error between the estimated depth and the ground truth through the hazing feature and depth feature learned in the backbone and decoders. We use MAE Loss to train the uncertainty, and the formula is as follows. 
\begin{equation} 
    \mathcal{L}_{MAE} = \frac{1}{n}\sum_i\vert \hat{e}_i - e_i\vert
    \label{eq:mae_loss}
\end{equation}
, where $e$ is $l1$ loss from estimated depth. To ensure that supervised learning does not affect depth estimation performance, all parameters except the UEB are not trained.

\begin{figure}[h!]
	\centering
	\def\width{1\columnwidth}%
    {%
		\includegraphics[width=\width]{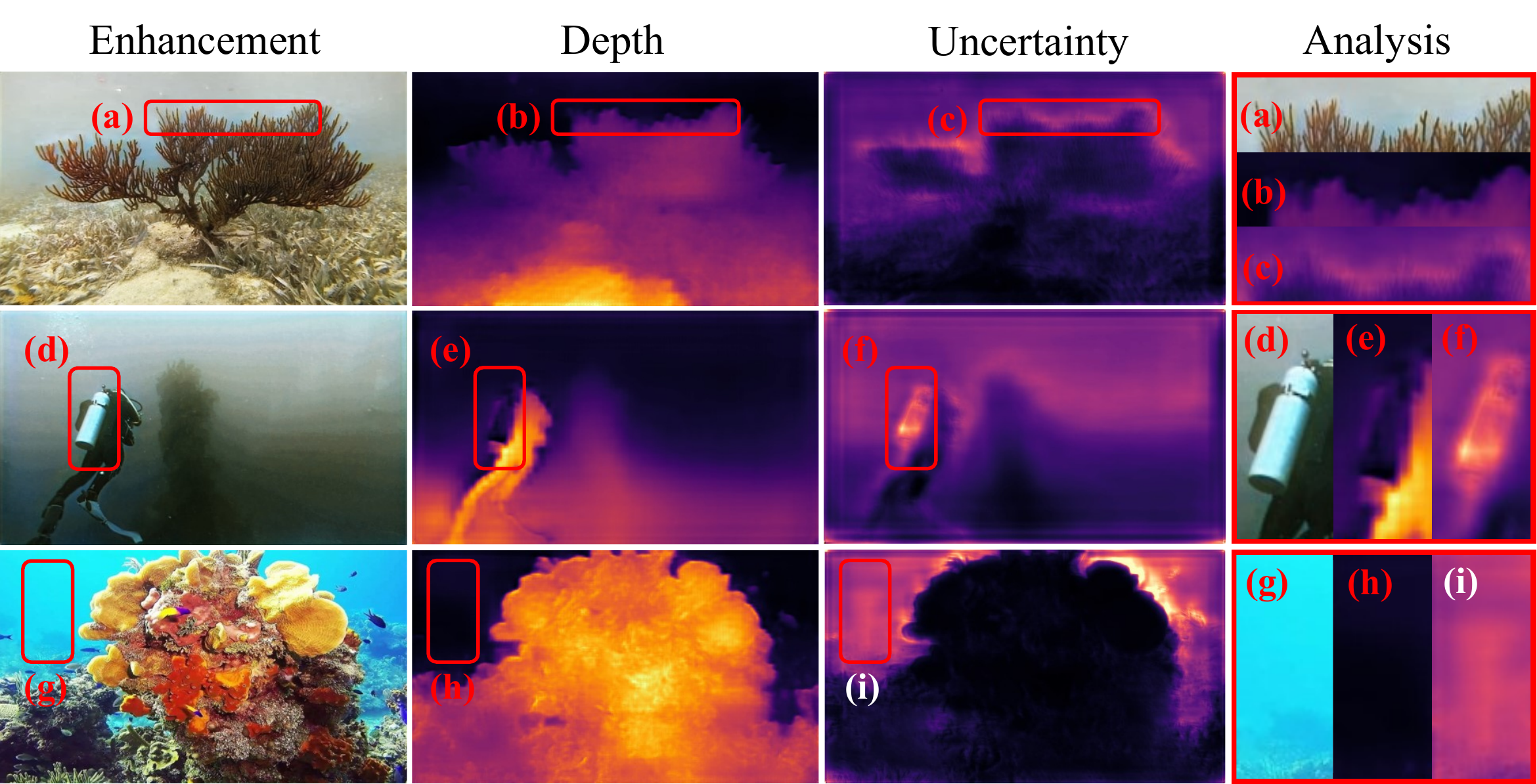}
	}
        \vspace{-3mm}
	\caption{Illustration of the analysis for depth uncertainty. This shows the enhanced image, depth map, and uncertainty for the three samples. And the red boxes in each $(a)$, $(b)$, $(c)$, $(d)$, $(e)$, $(f)$, $(g)$, $(h)$, and $(i)$ are used for qualitative comparison in the analysis figure.}
	\label{fig:uncertainty}
\end{figure}

With the above new supervised approach, we estimate the uncertainty simply but effectively. The uncertainty samples can be seen in \figref{fig:uncertainty}. In the first row, the edge area is bluish due to poor depth estimation. However, the edge in this area also has a high uncertainty. In the second row, the region where the depth prediction is unstable (red boxes) has high uncertainty. In the third row, the restoration starts to fail in the area with a large distance, but the depth uncertainty increases. This information can be used for 3-D reconstruction \cite{rahman2018sonar}, 3-D object localization \cite{chen2021efficient}, etc.

\begin{figure}[h]
	\centering
	\def\width{1\columnwidth}%
        {%
		\includegraphics[width=\width]{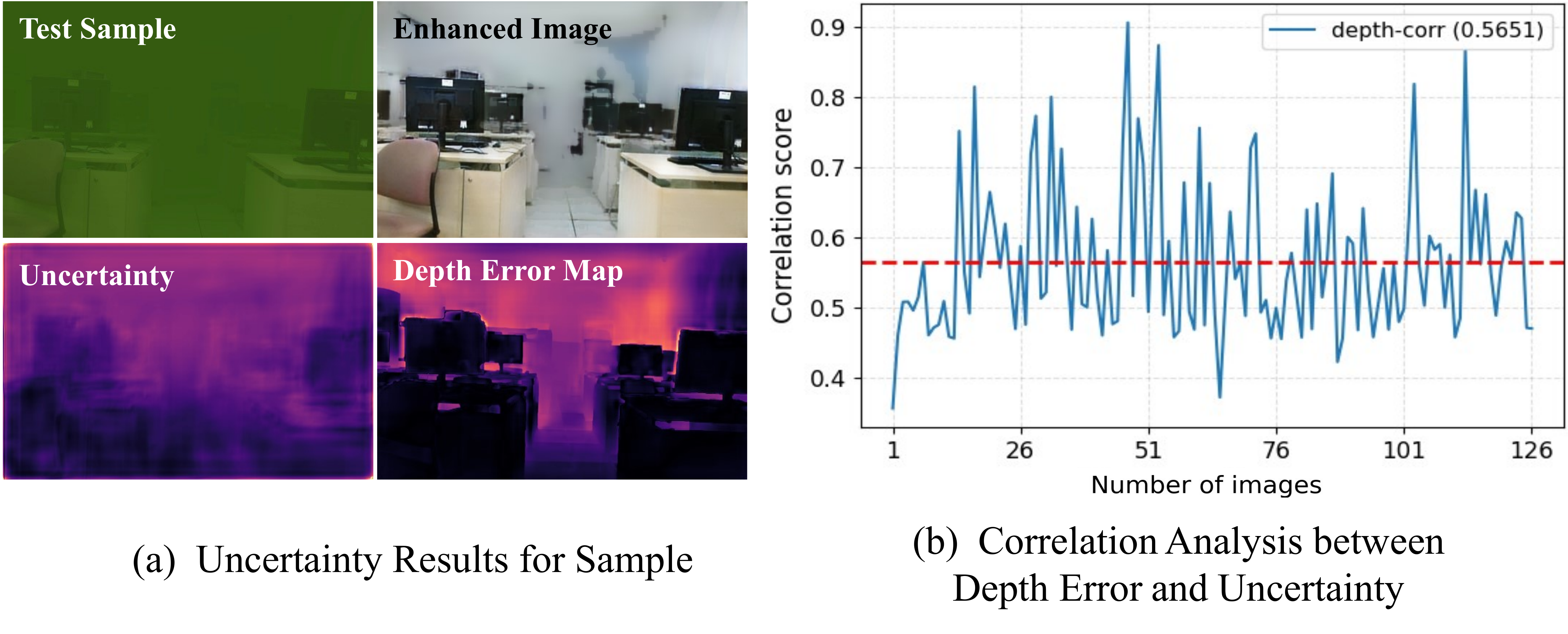}
	}
        \vspace{-3mm}
	\caption{Illustration of the quantitive analysis for depth uncertainty. This shows the test sample and the one type of graph: The graph is about correlation between estimated uncertainty and depth error map. The mean of correlations is a positive value of 0.5651.}
	\label{fig:uncertainty-correlation}
\end{figure}

\figref{fig:uncertainty-correlation} shows the results of the correlation analysis between uncertainty and depth error map for 126 hazed images in the Joint-ID test set. \figref{fig:uncertainty-correlation}(b) proves that the unreliability map has a positive correlation to the error map. This means that the estimated uncertainty and the depth errors have a positive proportional distribution.
\vspace{-2mm}
\subsection{End-to-End Training Objective}
In summary, the process for training the three tasks consists of two steps: $\lowercase\expandafter{\romannumeral1})$ jointly learning for supervised training of image enhancement and depth estimation; $\lowercase\expandafter{\romannumeral2})$ uncertainty estimation training through depth error. The training objective function for each step is as follows:
\begin{align} 
    \lowercase\expandafter{\romannumeral1}) &\qquad G_{joint}^* = \underset{G_{joint}}{\arg\min} \{ \underbrace{\lambda_{e}\cdot\mathcal{L}_{SSIM}}_{UIE}+\underbrace{\lambda_{d}\cdot\mathcal{L}_{SILog}}_{UMDE}\} \\ 
    \lowercase\expandafter{\romannumeral2}) &\qquad G_{uncert}^* = \underset{G_{uncert}}{\arg\min} \{ \lambda_{u}\cdot\mathcal{L}_{MAE} \}
    \label{eq:total_loss}
\end{align}
Here, $\lambda_\bigtriangleup$ symbols are scaling factors that represent the contributions of respective loss components. The selected values of $\lambda_e$, $\lambda_d$, $\lambda_u$ are 15.5, 1.5, and 512 in our implementation, respectively.

\vspace{-1mm}
\section{Experiments}
\subsection{Implementation Details} \label{Implementation}
As mentioned in \ref{section training overview}, TRIDENT training is supervised by paired data of the form $(\{X\}, \{E, D, U\})$ in Joint-ID's synthetic large training set. We used the Adam optimizer with a learning rate of $2\times10^{-4}$ and let TRIDENT converge with $40$ epochs and $32$ batch sizes. Our hardware system consists of a Nvidia RTX 3090 Ti graphics and a CPU 11th Intel i7 on Ubuntu 22.04. 
\begin{table*}[t!]
\caption{Quantitative performance comparison for image enhancement on PSNR, SSIM, Abs Rel, RMSE, and Sq Rel in reference datasets: scores are shown as $mean \pm \sqrt{variance}$; the first and second best scores (in each col) are colored {\rl{\textbf{red}}}, and {\bl{\textbf{blue}}}, respectively}
\label{tab:enhancement supervised}
\centering
\resizebox{0.975\textwidth}{!}{%

\begin{tabular}{c||lllll||lllll}
\toprule
                                 & \multicolumn{5}{c||}{\textbf{Joint-ID's Test set}}                                                                                                                                                                                                                                                                               & \multicolumn{5}{c}{\textbf{Test-L400}}                                                                                                                                                                                                                                                                                         \\ \cline{2-11} 
\multirow{-2}{*}{\textbf{Model}} & \multicolumn{1}{c|}{\textbf{PSNR $\uparrow$}}                                & \multicolumn{1}{c|}{\textbf{SSIM $\uparrow$}}                                 & \multicolumn{1}{c|}{\textbf{Abs Rel $\downarrow$}}                              & \multicolumn{1}{c|}{\textbf{RMSE $\downarrow$}}                              & \multicolumn{1}{c||}{\textbf{Sq Rel $\downarrow$}}          & \multicolumn{1}{c|}{\textbf{PSNR $\uparrow$}}                                & \multicolumn{1}{c|}{\textbf{SSIM $\uparrow$}}                                & \multicolumn{1}{c|}{\textbf{Abs Rel $\downarrow$}}                             & \multicolumn{1}{c|}{\textbf{RMSE $\downarrow$}}                               & \multicolumn{1}{c}{\textbf{Sq Rel $\downarrow$}}            \\ \hline \hline
DeepWave                         & \multicolumn{1}{l|}{18.47 $\pm$ 4.5}                                 & \multicolumn{1}{l|}{0.80 $\pm$ 0.15}                                  & \multicolumn{1}{l|}{58.40 $\pm$ 100}                                  & \multicolumn{1}{l|}{0.136 $\pm$ 0.07}                                 & \multicolumn{1}{l||}{15.5 $\pm$ 40.4}                                 & \multicolumn{1}{l|}{\bl{\textbf{20.32 $\pm$ 3.6}}} & \multicolumn{1}{l|}{\bl{\textbf{0.88 $\pm$ 0.07}}} & \multicolumn{1}{l|}{233.8 $\pm$ 270}                                 & \multicolumn{1}{l|}{\bl{\textbf{0.106 $\pm$ 0.051}}} & 23.5 $\pm$ 31.2                                 \\
FUnIE-GAN                        & \multicolumn{1}{l|}{17.21 $\pm$ 3.2}                                 & \multicolumn{1}{l|}{0.736 $\pm$ 0.13}                                 & \multicolumn{1}{l|}{86.26 $\pm$ 139}                                  & \multicolumn{1}{l|}{0.148 $\pm$ 0.06}                                 & \multicolumn{1}{l||}{31.0 $\pm$ 78.1}                       & \multicolumn{1}{l|}{16.92 $\pm$ 2.9}                                 & \multicolumn{1}{l|}{0.80 $\pm$ 0.08}                                 & \multicolumn{1}{l|}{317.1 $\pm$ 353}                                 & \multicolumn{1}{l|}{0.151 $\pm$ 0.059}                                 & 47.8 $\pm$ 79.2                                 \\
UGAN                             & \multicolumn{1}{l|}{18.73 $\pm$ 3.8}                                 & \multicolumn{1}{l|}{0.773 $\pm$ 0.15}                                 & \multicolumn{1}{l|}{71.24 $\pm$ 113}                                  & \multicolumn{1}{l|}{0.127 $\pm$ 0.06}                                 & \multicolumn{1}{l||}{20.5 $\pm$ 46.4}                      & \multicolumn{1}{l|}{18.28 $\pm$ 3.6}                                 & \multicolumn{1}{l|}{0.83 $\pm$ 0.10}                                 & \multicolumn{1}{l|}{262.2 $\pm$ 471}                                 & \multicolumn{1}{l|}{0.133 $\pm$ 0.061}                                 & 32.5 $\pm$ 76.9                                 \\
UIE-Dal                          & \multicolumn{1}{l|}{16.75 $\pm$ 4.8}                                 & \multicolumn{1}{l|}{0.792 $\pm$ 0.16}                                 & \multicolumn{1}{l|}{69.91 $\pm$ 156}                                  & \multicolumn{1}{l|}{0.171 $\pm$ 0.11}                                 & \multicolumn{1}{l||}{33.3 $\pm$ 121}                       & \multicolumn{1}{l|}{17.87 $\pm$ 3.8}                                 & \multicolumn{1}{l|}{0.86 $\pm$ 0.07} & \multicolumn{1}{l|}{316.8 $\pm$ 288}                                 & \multicolumn{1}{l|}{0.140 $\pm$ 0.062}                                 & 43.1 $\pm$ 55.3                                 \\
U-Shape                           & \multicolumn{1}{l|}{18.29 $\pm$ 3.9}                                 & \multicolumn{1}{l|}{0.751 $\pm$ 0.14}                                 & \multicolumn{1}{l|}{77.50 $\pm$ 123}                                  & \multicolumn{1}{l|}{0.135 $\pm$ 0.06}                                 & \multicolumn{1}{l||}{25.0 $\pm$ 58.0}                        & \multicolumn{1}{l|}{18.43 $\pm$ 3.5}                                 & \multicolumn{1}{l|}{0.80 $\pm$ 0.09}                                 & \multicolumn{1}{l|}{297.5 $\pm$ 281}                                 & \multicolumn{1}{l|}{0.130 $\pm$ 0.058}                                 & 44.4 $\pm$ 69.7                                 \\
UWCNN                            & \multicolumn{1}{l|}{13.44 $\pm$ 3.3}                                 & \multicolumn{1}{l|}{0.659 $\pm$ 0.17}                                 & \multicolumn{1}{l|}{89.65 $\pm$ 139}                                  & \multicolumn{1}{l|}{0.229 $\pm$ 0.08}                                 & \multicolumn{1}{l||}{32.5 $\pm$ 69.9}                        & \multicolumn{1}{l|}{14.73 $\pm$ 3.5}                                 & \multicolumn{1}{l|}{0.70 $\pm$ 0.08}                                 & \multicolumn{1}{l|}{519.0 $\pm$ 528}                                 & \multicolumn{1}{l|}{0.199 $\pm$ 0.082}                                 & 109.8 $\pm$ 170                               \\
Joint-ID                            & \multicolumn{1}{l|}{\bl{\textbf{20.46 $\pm$ 4.6}}} & \multicolumn{1}{l|}{\bl{\textbf{0.840 $\pm$ 0.15}}} & \multicolumn{1}{l|}{50.78 $\pm$ 100}   & \multicolumn{1}{l|}{\bl{\textbf{0.109 $\pm$ 0.06}}} & \multicolumn{1}{l||}{14.8 $\pm$ 46.0}         & \multicolumn{1}{l|}{\rl{\textbf{20.72 $\pm$ 3.9}}} & \multicolumn{1}{l|}{\rl{\textbf{0.90 $\pm$ 0.06}}}                                 & \multicolumn{1}{l|}{191.2 $\pm$ 234} & \multicolumn{1}{l|}{\rl{\textbf{0.102 $\pm$ 0.050}}} & 21.2 $\pm$ 52.5 \\
UIEC$^2$                            & \multicolumn{1}{l|}{19.76 $\pm$ 4.6} & \multicolumn{1}{l|}{0.811 $\pm$ 0.15} & \multicolumn{1}{l|}{\bl{\textbf{43.35 $\pm$ 92.1}}}   & \multicolumn{1}{l|}{0.117 $\pm$ 0.06} & \multicolumn{1}{l||}{\bl{\textbf{11.7 $\pm$ 41.6}}}         & \multicolumn{1}{l|}{19.75 $\pm$ 3.4} & \multicolumn{1}{l|}{0.85 $\pm$ 0.08}                                 & \multicolumn{1}{l|}{\bl{\textbf{174.7 $\pm$ 196}}} & \multicolumn{1}{l|}{0.111 $\pm$ 0.047} & \bl{\textbf{14.9 $\pm$ 25.0}} \\
\textbf{Ours}                    & \multicolumn{1}{l|}{\rl{\textbf{22.56 $\pm$ 4.8}}} & \multicolumn{1}{l|}{\rl{\textbf{0.851 $\pm$ 0.13}}} & \multicolumn{1}{l|}{\rl{\textbf{41.58 $\pm$ 86.5}}} & \multicolumn{1}{l|}{\rl{\textbf{0.084 $\pm$ 0.06}}} & \multicolumn{1}{l||}{\rl{\textbf{10.5 $\pm$ 38.4}}}    & \multicolumn{1}{l|}{18.68 $\pm$ 3.6}                                 & \multicolumn{1}{l|}{0.84 $\pm$ 0.10}                                 & \multicolumn{1}{l|}{\rl{\textbf{148.1 $\pm$ 201}}} & \multicolumn{1}{l|}{0.129 $\pm$ 0.065}                                 & \rl{\textbf{14.1 $\pm$ 33.2}} \\ \hline
\textbf{Ours ( \textless 5 )}                           & \multicolumn{1}{l|}{\textbf{23.82 $\pm$ 4.6}} & \multicolumn{1}{c|}{-} & \multicolumn{1}{l|}{\textbf{33.57 $\pm$ 66.2}}   & \multicolumn{1}{l|}{\textbf{0.074 $\pm$ 0.05}} & \multicolumn{1}{l||}{\textbf{5.84 $\pm$ 19.6}}         & \multicolumn{1}{l|}{\textbf{19.03 $\pm$ 4.2}} & \multicolumn{1}{c|}{-}  & \multicolumn{1}{l|}{211.2 $\pm$ 289} & \multicolumn{1}{l|}{\textbf{0.127 $\pm$ 0.050}} & 22.9 $\pm$ 83.1 \\
\textbf{Ours ( \textless 6 )}                            & \multicolumn{1}{l|}{23.24 $\pm$ 4.7} & \multicolumn{1}{c|}{-} & \multicolumn{1}{l|}{41.07 $\pm$ 111}   & \multicolumn{1}{l|}{0.080 $\pm$ 0.05} & \multicolumn{1}{l||}{9.16 $\pm$ 49.2}         & \multicolumn{1}{l|}{18.69 $\pm$ 3.9} & \multicolumn{1}{c|}{-} & \multicolumn{1}{l|}{163.3 $\pm$ 223} & \multicolumn{1}{l|}{0.129 $\pm$ 0.066} & {16.4 $\pm$ 49.2} \\
\textbf{Ours ( \textless 7 )}                    & \multicolumn{1}{l|}{23.06 $\pm$ 4.8} & \multicolumn{1}{c|}{-} & \multicolumn{1}{l|}{41.72 $\pm$ 88.9} & \multicolumn{1}{l|}{0.083 $\pm$ 0.05} & \multicolumn{1}{l||}{9.66 $\pm$ 36.9}    & \multicolumn{1}{l|}{18.61 $\pm$ 3.8}                                 & \multicolumn{1}{c|}{-}                                 & \multicolumn{1}{l|}{\textbf{155.6 $\pm$ 217}} & \multicolumn{1}{l|}{0.130 $\pm$ 0.065}                                 & {\textbf{15.4 $\pm$ 44.8}} \\
\bottomrule
\end{tabular}
}
\vspace{-2mm}
\end{table*}
\begin{table*}[t!]
\caption{Quantitative performance comparison for image enhancement on unsupervised image quality metrics, UIQM and UCIQE in real underwater datasets: scores are shown as $mean \pm \sqrt{variance}$; the first and second best scores (in each col) are colored {\rl{\textbf{red}}}, and {\bl{\textbf{blue}}}, respectively}
\label{tab:enhancement unsupervised}
\centering
\resizebox{1.0\textwidth}{!}{%

\begin{tabular}{c||ll||ll||ll||ll||cc}
\toprule
                                 & \multicolumn{2}{c||}{\textbf{UFO-120}}                                                                             & \multicolumn{2}{c||}{\textbf{U-90}}                                                                                & \multicolumn{2}{c||}{\textbf{USOD}}                                                                                & \multicolumn{2}{c||}{\textbf{USR-248}}                                                                             & \multicolumn{2}{c}{\textbf{Jamaica}}                                                                              \\ \cline{2-11} 
\multirow{-2}{*}{\textbf{Model}} & \multicolumn{1}{c|}{\textbf{UIQM $\uparrow$}}                                & \multicolumn{1}{c||}{\textbf{UCIQE $\uparrow$}}           & \multicolumn{1}{c|}{\textbf{UIQM $\uparrow$}}                                & \multicolumn{1}{c||}{\textbf{UCIQE $\uparrow$}}           & \multicolumn{1}{c|}{\textbf{UIQM $\uparrow$}}                                & \multicolumn{1}{c||}{\textbf{UCIQE $\uparrow$}}           & \multicolumn{1}{c|}{\textbf{UIQM $\uparrow$}}                                & \multicolumn{1}{c||}{\textbf{UCIQE $\uparrow$}}           & \multicolumn{1}{c|}{\textbf{UQIM $\uparrow$}}                                & \textbf{UCIQE $\uparrow$}  \\ \hline \hline
DeepWave                         & \multicolumn{1}{l|}{2.86 $\pm$ 0.48}                                 & 0.618 $\pm$ 0.05                                 & \multicolumn{1}{l|}{2.42 $\pm$ 0.52}                                 & 0.562 $\pm$ 0.06                                 & \multicolumn{1}{l|}{2.82 $\pm$ 0.45}                                 & 0.601 $\pm$ 0.05                                 & \multicolumn{1}{l|}{2.99 $\pm$ 0.44}                                 & 0.608 $\pm$ 0.04                                 & \multicolumn{1}{c|}{{\color[HTML]{FE0000} \textbf{3.00 $\pm$ 0.09}}} & 0.507 $\pm$ 0.02 \\
FUnIE-GAN                        & \multicolumn{1}{l|}{2.80 $\pm$ 0.48}                                 & 0.602 $\pm$ 0.06                                 & \multicolumn{1}{l|}{2.39 $\pm$ 0.58}                                 & 0.512 $\pm$ 0.08                                 & \multicolumn{1}{l|}{2.74 $\pm$ 0.51}                                 & 0.579 $\pm$ 0.08                                 & \multicolumn{1}{l|}{2.91 $\pm$ 0.42}                                 & 0.598 $\pm$ 0.05                                 & \multicolumn{1}{c|}{1.44 $\pm$ 0.09}                                 & 0.351 $\pm$ 0.03                                 \\
UGAN                             & \multicolumn{1}{l|}{2.77 $\pm$ 0.50}                                 & 0.607 $\pm$ 0.06                                 & \multicolumn{1}{l|}{2.33 $\pm$ 0.57}                                 & 0.498 $\pm$ 0.08                                 & \multicolumn{1}{l|}{2.70 $\pm$ 0.54}                                 & 0.578 $\pm$ 0.09                                 & \multicolumn{1}{l|}{2.89 $\pm$ 0.44}                                 & 0.603 $\pm$ 0.06                                 & \multicolumn{1}{c|}{1.37 $\pm$ 0.08}                                 & 0.344 $\pm$ 0.01                                 \\
UIE-Dal                          & \multicolumn{1}{l|}{2.85 $\pm$ 0.48}                                 & 0.627 $\pm$ 0.05                                 & \multicolumn{1}{l|}{2.19 $\pm$ 0.51}                                 & 0.533 $\pm$ 0.07                                 & \multicolumn{1}{l|}{2.86 $\pm$ 0.44}                                 & 0.609 $\pm$ 0.05                                 & \multicolumn{1}{l|}{2.98 $\pm$ 0.44}                                 & 0.620 $\pm$ 0.04                                 & \multicolumn{1}{c|}{2.58 $\pm$ 0.32}                                 & 0.414 $\pm$ 0.02                                 \\
U-Shape                           & \multicolumn{1}{l|}{2.69 $\pm$ 0.53}                                 & 0.622 $\pm$ 0.05                                 & \multicolumn{1}{l|}{\rl{\textbf{2.68 $\pm$ 0.48}}} & 0.581 $\pm$ 0.05                                 & \multicolumn{1}{l|}{2.70 $\pm$ 0.49}                                 & 0.605 $\pm$ 0.06                                 & \multicolumn{1}{l|}{2.81 $\pm$ 0.48}                                 & 0.614 $\pm$ 0.05                                 & \multicolumn{1}{c|}{1.73 $\pm$ 0.14}                                 & 0.389 $\pm$ 0.01                                 \\
UWCNN                            & \multicolumn{1}{l|}{\rl{\textbf{2.98 $\pm$ 0.42}}} & 0.599 $\pm$ 0.05                                 & \multicolumn{1}{l|}{2.23 $\pm$ 0.52}                                 & 0.508 $\pm$ 0.08                                 & \multicolumn{1}{l|}{\rl{\textbf{2.99 $\pm$ 0.35}}} & 0.579 $\pm$ 0.06                                 & \multicolumn{1}{l|}{3.04 $\pm$ 0.38}                                 & 0.595 $\pm$ 0.05                                 & \multicolumn{1}{c|}{1.44 $\pm$ 0.15}                                 & 0.333 $\pm$ 0.01                                 \\
Joint-ID                            & \multicolumn{1}{l|}{2.93 $\pm$ 0.40}                                 & 0.629 $\pm$ 0.05 & \multicolumn{1}{l|}{2.39 $\pm$ 0.56}                                 & 0.582 $\pm$ 0.05 & \multicolumn{1}{l|}{2.66 $\pm$ 0.15} & 0.610 $\pm$ 0.05 & \multicolumn{1}{l|}{3.03 $\pm$ 0.38} & 0.622 $\pm$ 0.04 & \multicolumn{1}{c|}{2.66 $\pm$ 0.15}                                 & \bl{\textbf{0.517 $\pm$ 0.03}}                                 \\
UIEC$^2$                            & \multicolumn{1}{l|}{2.95 $\pm$ 0.37}                                 & \bl{\textbf{0.632 $\pm$ 0.03}} & \multicolumn{1}{l|}{2.46 $\pm$ 0.50}                                 & \rl{\textbf{0.624 $\pm$ 0.03}} & \multicolumn{1}{l|}{\bl{\textbf{2.95 $\pm$ 0.36}}} & \rl{\textbf{0.631 $\pm$ 0.04}} & \multicolumn{1}{l|}{\rl{\textbf{3.07 $\pm$ 0.34}}} & \bl{\textbf{0.636 $\pm$ 0.03}} & \multicolumn{1}{c|}{2.76 $\pm$ 0.07}                                 & 0.485 $\pm$ 0.01                                 \\
\textbf{Ours}                    & \multicolumn{1}{l|}{\bl{\textbf{2.96 $\pm$ 0.39}}} & \rl{\textbf{0.644 $\pm$ 0.05}} & \multicolumn{1}{l|}{\bl{\textbf{2.47 $\pm$ 0.53}}} & \bl{\textbf{0.584 $\pm$ 0.04}} & \multicolumn{1}{l|}{2.92 $\pm$ 0.37}                                 & \bl{\textbf{0.615 $\pm$ 0.05}} & \multicolumn{1}{l|}{\bl{\textbf{3.05 $\pm$ 0.35}}} & \rl{\textbf{0.638 $\pm$ 0.05}} & \multicolumn{1}{c|}{\bl{\textbf{2.86 $\pm$ 0.18}}} & \rl{\textbf{0.541 $\pm$ 0.02}} \\ \bottomrule
\end{tabular}
}
\vspace{-2mm}
\end{table*}

\begin{figure*}[t!]
	\centering
	\def\width{2.0\columnwidth}%
    {%
		\includegraphics[width=\width]{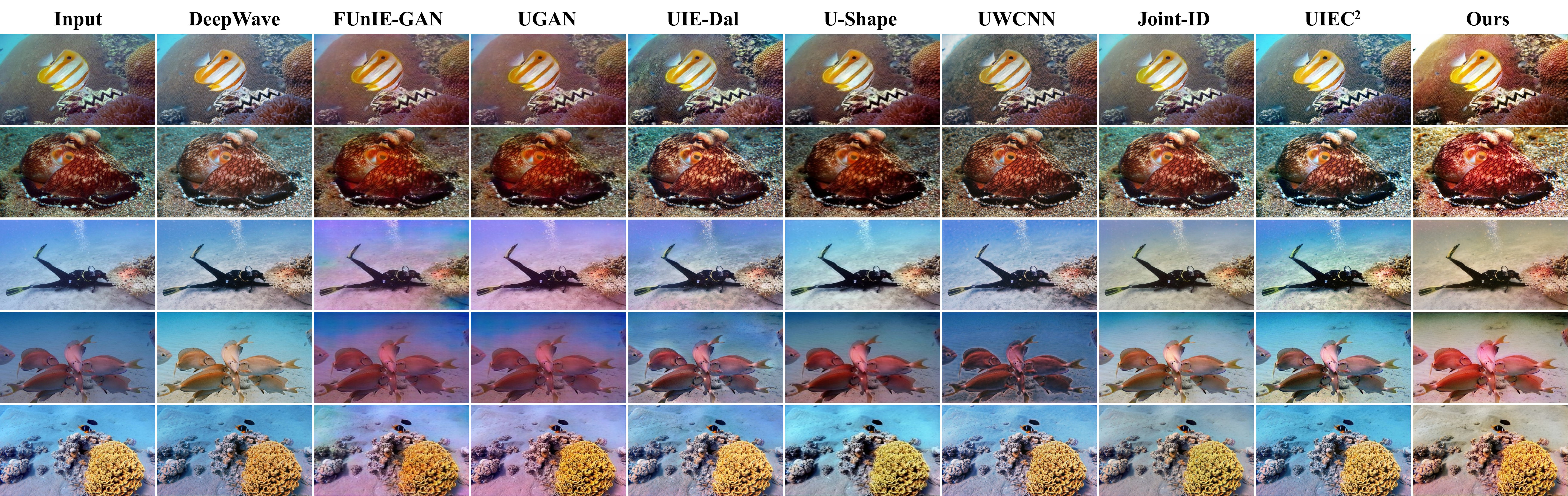}
	}
	\caption{Qualitative comparison of TRIDENT-enhanced images with SOTA models: DeepWave \cite{sharma2023wavelength}, FUnIE-GAN \cite{islam2020fast}, UGAN \cite{fabbri2018enhancing}, UIE-Dal \cite{uplavikar2019all}, U-Shape \cite{peng2023u}, UWCNN \cite{li2020underwater}, Joint-ID \cite{yang2023joint}, and UIEC$^2$ \cite{wang2021uiec}. Figures in rows 1-2 show a visual and qualitative comparison of colorfulness and brightness, while figures in rows 3-5 show the performance evaluation of depth-wise restoration.}
	\label{fig:total_enhancement}
\end{figure*}

\begin{figure}[t!]
	\centering
	\def\width{1.0\columnwidth}%
    {%
		\includegraphics[width=\width]{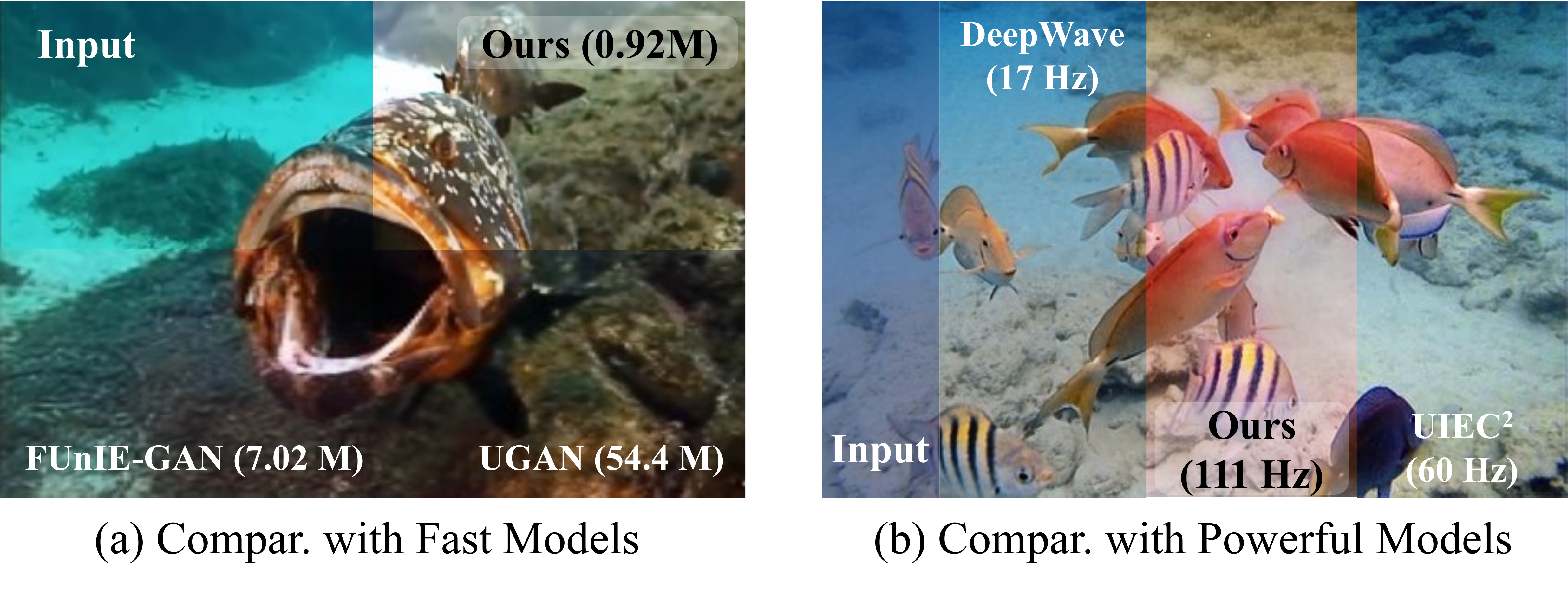}
	}
        \vspace{-3mm}
	\caption{Example results of perceptual restoration versus model memory and speed efficiency. The left shows the results of the fastest models, while the right shows the results of the models performing SOTA.}
	\label{fig:compare_fast_good}
\end{figure}

\vspace{-2mm}
\subsection{Evaluation: Image Enhancement}
First, the validation of \ac{UIE} consists of both quantitative and qualitative evaluations. Quantitative evaluation is based on the supervised metrics: peek signal-to-noise ratio (PSNR), structural similarity measure (SSIM), mean absolute relative error (Abs Rel), Root mean squared error (RMSE), and Squared relative error (Sq Rel), which follow the formulas as below:
\begin{itemize}
    \vspace{1mm}
    \item PSNR: $20 \cdot \log_{10}\left(\frac{255}{\sqrt{\frac{1}{n} \sum_{\hat{v} \in T_E} (v - \hat{v})^2}}\right)$
    \vspace{1mm}
    \item Abs Rel: ${\frac{1}{n}} \sum_{\hat{v} \in T_E} {\frac{{\vert \hat{v} - v \vert}}{v}}$
    \vspace{1mm}
    \item RMSE: $\sqrt{{\frac{1}{n}} \sum_{\hat{v} \in T_E} {\frac{\vert \hat{v} - v \vert}{v}}}$
    \vspace{1mm}
    \item Sq Rel: ${\frac{1}{n}} \sum_{\hat{v} \in T_E} {\frac{{\vert \hat{v} - v \vert}^2}{v}}$
    \vspace{2mm}
\end{itemize}
, where $T_E$ is the enhanced image, and $\hat{v}$ and $v$ are the pixel values belonging to the estimated enhanced image and ground truth, respectively. For unsupervised metrics, we used two metrics: UIQM \cite{panetta2015human} and UCIQE \cite{yang2015underwater}. The qualitative evaluation to validate the colorfulness and restoration of the whole scene is compared in \figref{fig:total_enhancement} and \figref{fig:compare_fast_good}. We compared the proposed method to recent well-known approaches; $\lowercase\expandafter{\romannumeral1})$ DeepWave, $\lowercase\expandafter{\romannumeral2})$ FUnIE-Gan, $\lowercase\expandafter{\romannumeral3})$ UGAN, $\lowercase\expandafter{\romannumeral4})$ UIE-Dal, $\lowercase\expandafter{\romannumeral5})$ U-shape, $\lowercase\expandafter{\romannumeral6})$ UWCNN, $\lowercase\expandafter{\romannumeral7})$ Joint-ID, $\lowercase\expandafter{\romannumeral8})$ UIEC$^2$. The following datasets were utilized for validation:
\begin{itemize}
    \item Joint-ID’s Test set \cite{yang2023joint}: A synthetic underwater dataset derived from the DIML RGB-D dataset. Unlike other datasets, it provides ground truth for both image enhancement and depth estimation simultaneously, facilitating quantitative evaluation for multi-task learning.
    \item L400 \cite{peng2023u}: A large-scale real underwater dataset named LUSI, proposed by \citet{peng2023u}, features a broad range of data distributions and provides pseudo-ground truth for clean images.
    \item UFO-120 \cite{islam2020simultaneous}: A dataset specifically designed for super-resolution tasks. The dataset includes ground truth for high-resolution and saliency maps.
    \item U-90 \cite{wang2021uiec}: A validation set of a comprehensive collection of underwater image enhancement benchmark (UIEB) datasets. It consists of 890 paired data, 90 of which are U-90.
    \item USOD \cite{islam2020svam}: The dataset is the first large-scale dataset developed specifically for underwater salient object detection.
    \item USR-248 \cite{islam2020underwater}: A dataset designed for underwater single image super-resolution tasks. It consists of high-resolution real underwater images and their corresponding low-resolution pairs.
    \item Jamaica \cite{li2017watergan}: The Jamaica dataset, also known as the Port Royal, Jamaica dataset, is part of a collection of underwater image datasets aimed at object detection.
    \vspace{2mm}
\end{itemize}

\begin{figure}[t!]
	\centering
	\def\width{1.0\columnwidth}%
    {%
		\includegraphics[width=\width]{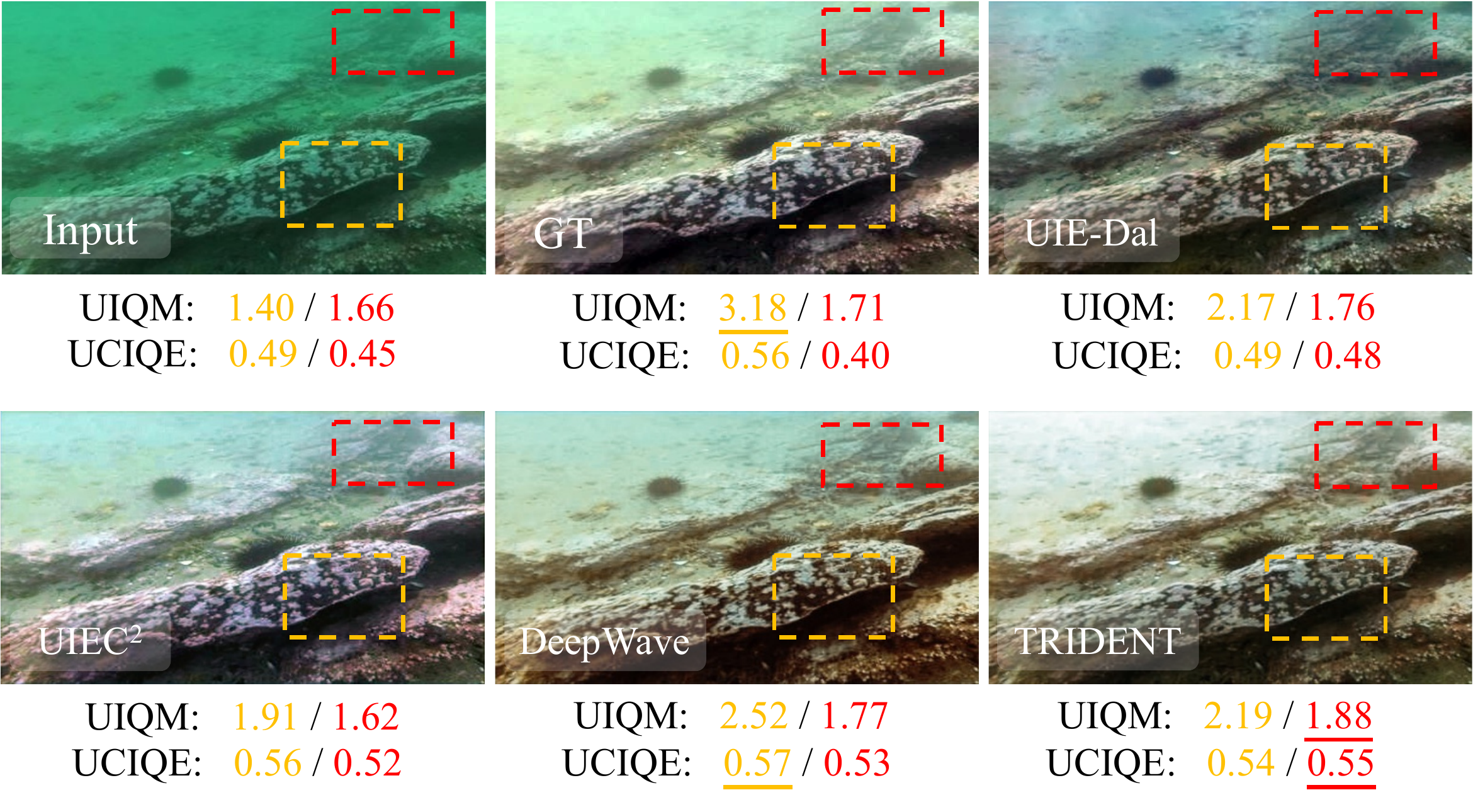}
	}
        \vspace{-5mm}
	\caption{A sample of results for image enhancement. Red boxes represent relatively distant parts of the scene, while yellow boxes represent relatively close parts. The quantitative results are from unsupervised metrics for each region. The highest values are \underline{underlined}.}
	\label{fig:compare_gt}
\vspace{-3mm}
\end{figure}

\begin{figure*}[t!]
\vspace{-2mm}
	\centering
	\def\width{2.025\columnwidth}%
        {%
            \includegraphics[width=\width]{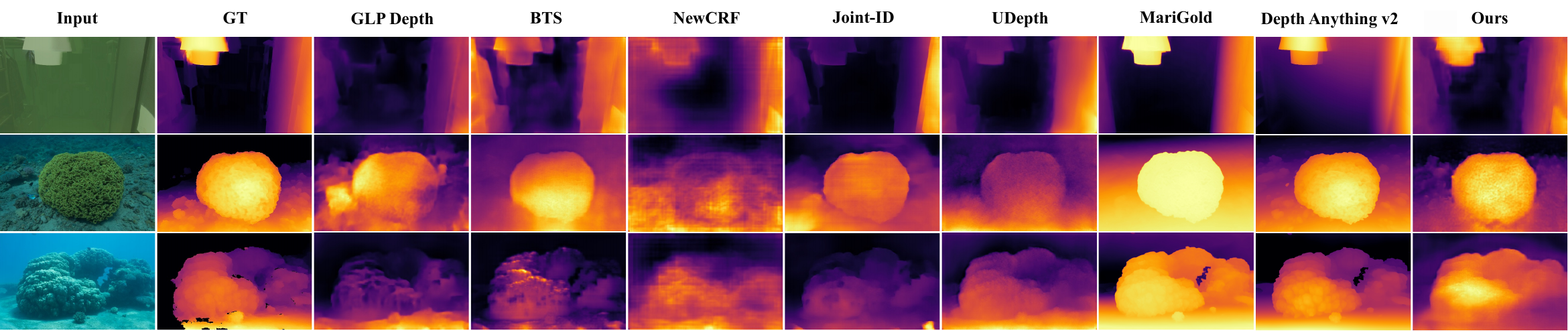}
	}
	\caption{Qualitative evaluation of depth estimation results to GLPDepth \cite{kim2022global}, BTS \cite{lee2019big}, NewCRF \cite{yuan2022neural}, UDepth \cite{yu2023udepth}, MariGold \cite{ke2024repurposing}, and Depth Anything V2 \cite{yang2024depth}. Figures in rows 1 show results for synthetic dataset, while figures in rows 2-3 show results for real underwater set.}
	\label{fig:depth estimation total}
\vspace{-2mm}
\end{figure*}

\subsubsection{Quantitative performance comparison} Supervised metrics are evaluated with ground truth, and unsupervised metrics are evaluated without ground truth. The supervised-metric evaluations were performed on Joint-ID's Test-set \cite{yang2023joint} and L400 \cite{peng2023u}. As presented in Table \ref{tab:enhancement supervised}, our model outperforms other models on the joint-ID test set, which exhibits severe hazing conditions. It also attains the best performance for the metrics Abs Rel and Sq Rel on the pseudo-reference L400 test set. However, the results indicate that ours does not perform best on PSNR, SSIM, and RMSE. This is because the ground truth is not a fully undistorted image; specifically, distant regions are not properly calibrated. In \figref{fig:compare_gt}, other models, including the ground truth, show a decline in performance with increasing distance on the L400 test set. Additionally, our model performs best at larger distances for the unsupervised metrics UIQM \cite{panetta2015human} and UCIQE \cite{yang2015underwater}. This explains why we did not achieve a high score on the L400 test set. The bottom three rows of Table \ref{tab:enhancement supervised} show the quantitative evaluation results after filtering out the unreliable parts of the predicted intensities through the depth uncertainty. $< value$ means the uncertainty threshold, with many pixels filtered in the order $<5$, $<6$, $<7$.

 As demonstrated in Table \ref{tab:enhancement unsupervised}, for unsupervised-metric evaluation with some reference sets (UFO-120 \cite{islam2020simultaneous}, U-90 \cite{wang2021uiec}, USOD \cite{islam2020svam}, USR-248 \cite{islam2020underwater}, Jamaica \cite{li2017watergan}), which evaluates edge sharpness, and color contrast, ours and UIEC$^2$ consistently show the best results for all real underwater test sets. These results indicate that our approach exhibits high color contrast, which can positively influence tasks such as feature matching and saliency detection in the fields of computer vision and robotics. The application proof of this is proved in more detail in section \ref{Application}.

The following sections \ref{sec: speed and memory} and \ref{sec: method analysis} show the quantitative evaluation of image enhancement and depth estimation, qualitative analysis, and experimental results on the model efficiency. To ensure fair evaluation, the models were trained on the Joint-ID training set.

\subsubsection{Qualitative performance comparison} Next, we qualitatively compared the perceptual restoration of our model with other models. In \figref{fig:total_enhancement}, from the first to second rows, our enhanced images outperform others in colorfulness and color contrast. The results in the third to fifth rows represent the performance of our model on the distant regions of the images. \figref{fig:compare_fast_good} represents direct comparisons of our model and recent approaches on two categories: efficiency and performance. \figref{fig:compare_fast_good}(a) shows the comparison results for color correction, edge details, and model size between our model and the efficiency-targeted models. It also shows that ours has the best enhancement results even with the lowest memory size. In \figref{fig:compare_fast_good}(b), the qualitative and speed comparisons with \ac{SOTA} models are shown, and it is evident that only ours shows perceptual enhancement regardless of distance, and the speed is the fastest at 111Hz.

This means that the proposed method achieves efficiency with robust restoration. Specific speed and memory comparisons are shown in Table \ref{tab:compare memory speed}.

\vspace{-3mm}
\subsection{Evaluation: Depth Estimation}
\begin{table*}[t!]
\caption{Quantitative performance comparison for depth estimation on Silog, Abs Rel, and Sq Rel in our reference datasets: scores are shown as $mean \pm \sqrt{variance}$; the first, second and third best scores (in each col) are colored {\color[HTML]{FE0000} \textbf{red}}, \bl{\textbf{blue}} and \underline{underline}, respectively}
\label{tab:depth estimation supervised}
\centering
\resizebox{1.0\textwidth}{!}{%

\begin{tabular}{c||ccc||ccc||ccc}
\toprule
                                 & \multicolumn{3}{c||}{\textbf{Joint-ID's Test set}}                                                                                                                                                      & \multicolumn{3}{c||}{\textbf{Sea-thru}}                                                                                                                                                   & \multicolumn{3}{c}{\textbf{SQUID}}                                                                                                                                                     \\ \cline{2-10} 
\multirow{-2}{*}{\textbf{Model}} & \multicolumn{1}{c|}{\textbf{Silog $\downarrow$}}                                & \multicolumn{1}{c|}{\textbf{Abs Rel $\downarrow$}}                              & \textbf{Sq Rel $\downarrow$}                               & \multicolumn{1}{c|}{\textbf{Silog $\downarrow$}}                                & \multicolumn{1}{c|}{\textbf{Abs Rel $\downarrow$}}                              & \textbf{Sq Rel $\downarrow$}                               & \multicolumn{1}{c|}{\textbf{Silog $\downarrow$}}                                & \multicolumn{1}{c|}{\textbf{Abs Rel $\downarrow$}}                              & \textbf{Sq Rel $\downarrow$}                              \\ \hline \hline
BTS                     & \multicolumn{1}{c|}{\underline{16.45 $\pm$ 6.75}} & \multicolumn{1}{c|}{\underline{0.206 $\pm$ 0.17}} & 0.189 $\pm$ 0.32 & \multicolumn{1}{c|}{17.60 $\pm$ 5.57} & \multicolumn{1}{c|}{\underline{0.231 $\pm$ 0.06}} & 0.198 $\pm$ 0.16 & \multicolumn{1}{c|}{27.76 $\pm$ 13.2} & \multicolumn{1}{c|}{0.689 $\pm$ 0.18}                                 & 4.50 $\pm$ 2.79                                 \\
GLPDepth                & \multicolumn{1}{c|}{18.49 $\pm$ 8.03}                                 & \multicolumn{1}{c|}{0.247 $\pm$ 0.20}                                 & 0.244 $\pm$ 0.35                                 & \multicolumn{1}{c|}{21.20 $\pm$ 6.41}                                 & \multicolumn{1}{c|}{0.248 $\pm$ 0.17}                                 & 0.219 $\pm$ 0.29                                 & \multicolumn{1}{c|}{30.13 $\pm$ 14.5}                                 & \multicolumn{1}{c|}{0.643 $\pm$ 0.20} & 4.30 $\pm$ 2.90                                 \\
NewCRF                  & \multicolumn{1}{c|}{22.99 $\pm$ 9.26}                                 & \multicolumn{1}{c|}{0.314 $\pm$ 0.25}                                 & 0.398 $\pm$ 0.54                                 & \multicolumn{1}{c|}{20.28 $\pm$ 6.49}                                 & \multicolumn{1}{c|}{0.246 $\pm$ 0.16} & 0.228 $\pm$ 0.40                                 & \multicolumn{1}{c|}{34.24 $\pm$ 16.7}                                 & \multicolumn{1}{c|}{0.648 $\pm$ 0.18}                                 & 4.12 $\pm$ 2.67 \\
Joint-ID                  & \multicolumn{1}{c|}{\bl{\textbf{14.94 $\pm$ 8.18}}}                                 & \multicolumn{1}{c|}{\bl{\textbf{0.170 $\pm$ 0.22}}}                                 & \bl{\textbf{0.156 $\pm$ 0.48}}                                & \multicolumn{1}{c|}{21.06 $\pm$ 7.99}                                 & \multicolumn{1}{c|}{0.259 $\pm$ 0.17}                                 & 0.224 $\pm$ 0.25                                & \multicolumn{1}{c|}{31.26 $\pm$ 14.5} & \multicolumn{1}{c|}{0.530 $\pm$ 0.21}                                 & 3.34 $\pm$ 2.06                                 \\
UDepth                  & \multicolumn{1}{c|}{18.30 $\pm$ 7.32}                                 & \multicolumn{1}{c|}{0.238 $\pm$ 0.18}                                 & 0.226 $\pm$ 0.31                                 & \multicolumn{1}{c|}{17.78 $\pm$ 5.31}                                 & \multicolumn{1}{c|}{0.268 $\pm$ 0.18}                                 & 0.203 $\pm$ 0.20                                 & \multicolumn{1}{c|}{\underline{24.94 $\pm$ 13.0}} & \multicolumn{1}{c|}{0.687 $\pm$ 0.20}                                 & 4.63 $\pm$ 2.96                                 \\
MariGold               & \multicolumn{1}{c|}{18.86 $\pm$ 8.31}                                 & \multicolumn{1}{c|}{0.244 $\pm$ 0.18}                                 & 0.196 $\pm$ 0.38                                 & \multicolumn{1}{c|}{\bl{\textbf{16.59 $\pm$ 4.91}}}                                 & \multicolumn{1}{c|}{\bl{\textbf{0.205 $\pm$ 0.08}}}                                & \bl{\textbf{0.181 $\pm$ 0.13}}                                 & \multicolumn{1}{c|}{\bl{\textbf{21.32 $\pm$ 12.1}}} & \multicolumn{1}{c|}{\bl{\textbf{0.411 $\pm$ 0.16}}}                                 & \underline{2.52 $\pm$ 2.46}                                 \\
Depth Anything v2       & \multicolumn{1}{c|}{17.13 $\pm$ 7.88}                                 & \multicolumn{1}{c|}{0.243 $\pm$ 0.19}                                 & \underline{0.178 $\pm$ 0.21}                                 & \multicolumn{1}{c|}{{\color[HTML]{FE0000} \textbf{14.71 $\pm$ 4.78}}}                                 & \multicolumn{1}{c|}{{\color[HTML]{FE0000} \textbf{0.183 $\pm$ 0.07}}}                                 & {\color[HTML]{FE0000} \textbf{0.173 $\pm$ 0.12}}                                 & \multicolumn{1}{c|}{{\color[HTML]{FE0000} \textbf{19.14 $\pm$ 10.2}}} & \multicolumn{1}{c|}{{\color[HTML]{FE0000} \textbf{0.382 $\pm$ 0.14}}}                                 & \bl{\textbf{2.21 $\pm$ 1.98}}                                 \\
\textbf{Ours}                    & \multicolumn{1}{c|}{{\color[HTML]{FE0000} \textbf{12.35 $\pm$ 5.21}}} & \multicolumn{1}{c|}{{\color[HTML]{FE0000} \textbf{0.136 $\pm$ 0.11}}} & {\color[HTML]{FE0000} \textbf{0.091 $\pm$ 0.09}} & \multicolumn{1}{c|}{\underline{17.17 $\pm$ 5.01}} & \multicolumn{1}{c|}{0.248 $\pm$ 0.17}                                 & {\underline{0.196 $\pm$ 0.18}} & \multicolumn{1}{c|}{33.43 $\pm$ 15.1}                                 & \multicolumn{1}{c|}{\underline{0.481 $\pm$ 0.21}} & {\color[HTML]{FE0000} \textbf{2.07 $\pm$ 1.78}} \\ \hline
\textbf{Ours ( \textless 5 )}    & \multicolumn{1}{c|}{\textbf{11.98 $\pm$ 5.09}}                        & \multicolumn{1}{c|}{\textbf{0.138 $\pm$ 0.08}}                        & \textbf{0.073 $\pm$ 0.07}                        & \multicolumn{1}{c|}{17.22 $\pm$ 4.86}                                 & \multicolumn{1}{c|}{\textbf{0.243 $\pm$ 0.14}}                        & \textbf{0.194 $\pm$ 0.19}                        & \multicolumn{1}{c|}{\textbf{32.78 $\pm$ 14.7}}                        & \multicolumn{1}{c|}{\textbf{0.476 $\pm$ 0.24}}                        & \textbf{1.97 $\pm$ 1.62}                        \\
\textbf{Ours ( \textless 6 )}    & \multicolumn{1}{c|}{12.15 $\pm$ 5.12}                                 & \multicolumn{1}{c|}{0.140 $\pm$ 0.08}                                 & 0.076 $\pm$ 0.07                                 & \multicolumn{1}{c|}{\textbf{17.19 $\pm$ 4.60}}                        & \multicolumn{1}{c|}{0.244 $\pm$ 0.14}                                 & 0.196 $\pm$ 0.19                                 & \multicolumn{1}{c|}{32.90 $\pm$ 15.1}                                 & \multicolumn{1}{c|}{0.488 $\pm$ 0.24}                                 & 2.20 $\pm$ 1.78                                 \\
\textbf{Ours ( \textless 7 )}    & \multicolumn{1}{c|}{12.26 $\pm$ 5.28}                               & \multicolumn{1}{c|}{0.141 $\pm$ 0.09}                                 & 0.079 $\pm$ 0.08                                 & \multicolumn{1}{c|}{17.20 $\pm$ 4.54}                        & \multicolumn{1}{c|}{0.247 $\pm$ 0.16}                                 & 0.196 $\pm$ 0.19                                 & \multicolumn{1}{c|}{32.93 $\pm$ 15.0}                                 & \multicolumn{1}{c|}{0.490 $\pm$ 0.26}                                 & 2.20 $\pm$ 1.79                                \\    
\bottomrule
\end{tabular}
}
\vspace{-2mm}
\end{table*}
The quantitative evaluation of the comparison for depth estimation was performed only with supervised metrics. The metrics include Abs Rel ($T_E \rightarrow T_D$, $\hat{v} \rightarrow \hat{d}$, $v \rightarrow d$) and Sq Rel ($T_E \rightarrow T_D$, $\hat{v} \rightarrow \hat{d}$, $v \rightarrow d$), which are the same equations in Silog and \ac{UIE} evaluation; $T_D$ is the depth image, and $\hat{d}$, $d$ are ground truth and predicted depth of one pixel at the same index, respectively. The comparison \ac{SOTA} models are as follows: $\lowercase\expandafter{\romannumeral1})$ BTS, $\lowercase\expandafter{\romannumeral2})$ GLPDepth, $\lowercase\expandafter{\romannumeral3})$ NewCRF, $\lowercase\expandafter{\romannumeral4})$ Joint-ID, $\lowercase\expandafter{\romannumeral5})$ UDepth, $\lowercase\expandafter{\romannumeral6})$ MariGold, and $\lowercase\expandafter{\romannumeral7})$ Depth Anything V2. The qualitative evaluation was performed on synthetic datasets and several real underwater scenes:
\begin{itemize}
    \item Sea-thru \cite{akkaynak2019sea}: The Sea-thru dataset is a collection of underwater images designed to facilitate algorithm development and validation. It is real underwater and provides dense depth ground truth.
    \item SQUID \cite{berman2020underwater}: The SQUID (Stereo Quantitative Underwater Image Dataset) is a specialized dataset created to support research in underwater 3-D reconstruction. To get distance maps, the 3-D structure of the scenes was calculated using stereo imaging techniques.
    \vspace{2mm}
\end{itemize}

\subsubsection{Quantitative performance comparison} \label{depth quatitative performance comparision}
The supervised metrics for quantitative evaluation were processed through Joint-ID's Test set, Sea-thru \cite{akkaynak2019sea}, and the SQUID \cite{berman2020underwater} Test set, and the results are shown in Table \ref{tab:enhancement unsupervised}. Our approach outperforms Joint-ID on its test set and demonstrates strong overall performance on the Sea-thru and SQUID datasets. However, large-scale models, such as MariGold and Depth Anything V2, also performed well on underwater depth estimation. Despite this, as evidenced by the results in Tables \ref{tab:depth estimation supervised} and \ref{tab:compare memory speed}, our method excels in \ac{UIE} and \ac{UMDE}, while also being efficient in terms of memory usage and processing speed. Finally, it can be seen in the bottom three lines of Table \ref{tab:depth estimation supervised} that uncertainty filtering improves the overall depth accuracy.

\subsubsection{Qualitative performance comparison} The visual comparisons presented in \figref{fig:depth estimation total} support our analysis, demonstrating that our method generally exhibits higher accuracy and enhanced edge sharpness compared to other models, with the exception of MariGold and Depth Anything V2. Notably, while most other models struggle with accurately interpreting spatial relationships between objects, particularly in severely hazed regions (e.g., the first row), our approach excels in understanding depth relationships. Although MariGold and Depth Anything V2 produce strong results, as discussed in section \ref{depth quatitative performance comparision}, they are not applicable to real-time applications.

\subsection{Evaluation: Speed and Memory} \label{sec: speed and memory}
\begin{table}[h!]
\caption{Quantitative performance comparison for model efficiency on memory size and inference time: the first and second best scores (in each col) are colored {\color[HTML]{FE0000} \textbf{red}} and \bl{\textbf{blue}}, respectively. FPS (a), FPS (b) are mean using GPU and CPU, respectively.}
\label{tab:compare memory speed}
\centering
\resizebox{0.5\textwidth}{!}{%

\begin{tabular}{ccccccc}
\toprule
\multicolumn{1}{c||}{\textbf{Method}}             & \multicolumn{1}{c|}{\textbf{\# Tasks}} & \multicolumn{1}{c|}{\textbf{\# Params}}                     & \multicolumn{1}{c|}{\textbf{Memory}}                          & \multicolumn{1}{c|}{\textbf{FPS (a)}}                                            & \multicolumn{1}{c|}{\textbf{FPS (b)}}                                             & \textbf{Output Size}                                                             \\ \midrule \hline
\multicolumn{7}{c}{\textbf{Enhancement}}     

\\ \hline 
\multicolumn{1}{c||}{\textbf{DeepWave}}           & \multicolumn{1}{c|}{2}                 & \multicolumn{1}{c|}{{\color[HTML]{FE0000} \textbf{0.28 M}}} & \multicolumn{1}{c|}{{\color[HTML]{FE0000} \textbf{1.2 MB}}}   & \multicolumn{1}{c|}{17 Hz}                                  & \multicolumn{1}{c|}{3.2 Hz}                                  & (3, 256, 256)                                                                    \\
\multicolumn{1}{c||}{\textbf{FUnIE-GAN}}          & \multicolumn{1}{c|}{1}                 & \multicolumn{1}{c|}{7.02 M}                                 & \multicolumn{1}{c|}{28.1 MB}                                  & \multicolumn{1}{c|}{{\color[HTML]{FE0000} \textbf{417 Hz}}} & \multicolumn{1}{c|}{{\color[HTML]{FE0000} \textbf{32.4 Hz}}} & (3, 256, 256)                                                                    \\
\multicolumn{1}{c||}{\textbf{UGAN}}               & \multicolumn{1}{c|}{1}                 & \multicolumn{1}{c|}{54.4 M}                                 & \multicolumn{1}{c|}{217.7 MB}                                 & \multicolumn{1}{c|}{\bl{\textbf{330 Hz}}} & \multicolumn{1}{c|}{\bl{\textbf{15.4 Hz}}} & (3, 256, 256)                                                                    \\
\multicolumn{1}{c||}{\textbf{UIE-Dal}}            & \multicolumn{1}{c|}{1}                 & \multicolumn{1}{c|}{13.4 M}                                 & \multicolumn{1}{c|}{53.7 MB}                                  & \multicolumn{1}{c|}{211 Hz}                                 & \multicolumn{1}{c|}{8.2 Hz}                                  & (3, 256, 256)                                                                    \\
\multicolumn{1}{c||}{\textbf{UIEC2}}              & \multicolumn{1}{c|}{1}                 & \multicolumn{1}{c|}{\bl{\textbf{0.55 M}}} & \multicolumn{1}{c|}{\bl{\textbf{2.2 MB}}}   & \multicolumn{1}{c|}{60 Hz}                                  & \multicolumn{1}{c|}{3.7 Hz}                                  & (3, 256, 256)                                                                    \\
\multicolumn{1}{c||}{\textbf{U-shape}}            & \multicolumn{1}{c|}{1}                 & \multicolumn{1}{c|}{31.6 M}                                 & \multicolumn{1}{c|}{126.6 MB}                                 & \multicolumn{1}{c|}{38 Hz}                                  & \multicolumn{1}{c|}{4.3 Hz}                                  & (3, 256, 256)                                                                    \\ 

\midrule \hline
\multicolumn{7}{c}{\textbf{Depth Estimation}} 

\\ \hline
\multicolumn{1}{c||}{\textbf{BTS}}                & \multicolumn{1}{c|}{1}                 & \multicolumn{1}{c|}{113 M}                                  & \multicolumn{1}{c|}{452.2 MB}                                 & \multicolumn{1}{c|}{\bl{\textbf{67 Hz}}}  & \multicolumn{1}{c|}{4.5 Hz}                                  & (1, 256, 256)                                                                    \\
\multicolumn{1}{c||}{\textbf{GLPDepth}}           & \multicolumn{1}{c|}{1}                 & \multicolumn{1}{c|}{\bl{\textbf{61.2 M}}} & \multicolumn{1}{c|}{\bl{\textbf{245.2 MB}}} & \multicolumn{1}{c|}{47 Hz}                                  & \multicolumn{1}{c|}{\bl{\textbf{5.8 Hz}}}  & (1, 256, 256)                                                                    \\
\multicolumn{1}{c||}{\textbf{NewCRF}}             & \multicolumn{1}{c|}{1}                 & \multicolumn{1}{c|}{270 M}                                  & \multicolumn{1}{c|}{1082 MB}                                  & \multicolumn{1}{c|}{56 Hz}                                  & \multicolumn{1}{c|}{1.2 Hz}                                  & (1, 256, 256)                                                                    \\
\multicolumn{1}{c||}{\textbf{UDepth}}             & \multicolumn{1}{c|}{1}                 & \multicolumn{1}{c|}{{\color[HTML]{FE0000} \textbf{15.6 M}}} & \multicolumn{1}{c|}{{\color[HTML]{FE0000} \textbf{62.7 MB}}}  & \multicolumn{1}{c|}{{\color[HTML]{FE0000} \textbf{135 Hz}}} & \multicolumn{1}{c|}{{\color[HTML]{FE0000} \textbf{25.1 Hz}}} & (1, 128, 128)                                                                    \\ 
\multicolumn{1}{c||}{\textbf{MariGold}}             & \multicolumn{1}{c|}{1}                 & \multicolumn{1}{c|}{1290 M} & \multicolumn{1}{c|}{5157 MB}  & \multicolumn{1}{c|}{0.052 Hz} & \multicolumn{1}{c|}{ - Hz} & (1, 256, 256)                                                          \\ 
\multicolumn{1}{c||}{\textbf{Depth Anything v2}}             & \multicolumn{1}{c|}{1}                 & \multicolumn{1}{c|}{335.4 M} & \multicolumn{1}{c|}{1341 MB}  & \multicolumn{1}{c|}{2.8 Hz} & \multicolumn{1}{c|}{0.095 Hz} & (1, 256, 256)                                            \\ 

\midrule \hline
\multicolumn{7}{c}{\textbf{Enhancement + Depth Estimation (+ Uncertainty)}}  
\\ \hline 
\multicolumn{1}{c||}{\textbf{Joint-ID}}               & \multicolumn{1}{c|}{2}        & \multicolumn{1}{c|}{150 M}                        & \multicolumn{1}{c|}{601.9 MB}                          & \multicolumn{1}{c|}{15.3 Hz}                        & \multicolumn{1}{c|}{3.1 Hz}                        & \begin{tabular}[c]{@{}c@{}}(3, 256, 256), \\ (1, 256, 256)\end{tabular} \\
\multicolumn{1}{c||}{\textbf{Ours}}               & \multicolumn{1}{c|}{\textbf{2}}        & \multicolumn{1}{c|}{\textbf{0.92 M}}                        & \multicolumn{1}{c|}{\textbf{3.9 MB}}                          & \multicolumn{1}{c|}{\textbf{111 Hz}}                        & \multicolumn{1}{c|}{\textbf{18.7 Hz}}                        & \textbf{\begin{tabular}[c]{@{}c@{}}(3, 256, 256), \\ (1, 256, 256)\end{tabular}} \\
\multicolumn{1}{c||}{\textbf{\begin{tabular}[c]{@{}c@{}}Ours \\ \footnotesize{(+Uncerainty)}\end{tabular}}} & \multicolumn{1}{c|}{\textbf{3}}        & \multicolumn{1}{c|}{\textbf{0.96 M}}                        & \multicolumn{1}{c|}{\textbf{4.1 MB}}                          & \multicolumn{1}{c|}{\textbf{102 Hz}}                        & \multicolumn{1}{c|}{\textbf{16.5 Hz}}                        & \textbf{+ (1, 256, 256)}                              \\    
\bottomrule
\end{tabular}
}
\end{table}
We also compare the computational efficiency of all models for \ac{UIE} and \ac{UMDE} in Table \ref{tab:compare memory speed}. Among the other models for \ac{UIE}, DeepWave and UIEC$^2$ have the smallest model size, which has good quantitative metrics. But, compared to ours, they are significantly slower, by about $670\%$ and $191\%$ in terms of $FPS(a)$, respectively. It proves that ours is a robot-friendly fast model. Meanwhile, the fastest models, FUnIE-GAN and UGAN, effectively show fast results on inference time. But, their qualitative results are not good, as shown in \figref{fig:total_enhancement} and \figref{fig:compare_fast_good}, and their sizes are $7.02M$ and $54.4M$, respectively, which means they are about $7.6\times$ and $59.1\times$ heavier than ours. Our efficiency is more pronounced when compared to \ac{UMDE} models. The smallest comparison models are UDepth and GLPDepth, which are about $17\times$ and $66\times$ larger than ours, respectively. Furthermore, our model demonstrates significant advantages in inference time. Excluding UDepth, our model achieves speed improvements ranging from $3.32\times$ to $16.08\times$ in $FPS(b)$ compared to other methods. It’s worth noting that while UDepth approaches similar speeds, it suffers from a critical drawback, producing output at only half the original resolution. These results underscore the efficiency of our model in balancing lightweight architecture with rapid inference time. Most importantly, TRIDENT achieves computational and memory efficiency while simultaneously performing two or three tasks. This capability highlights the feasibility of executing multiple tasks swiftly, even in complex environments.

\subsection{Evaluation: Method Analysis} \label{sec: method analysis}
\subsubsection{Ablation Study for Three Stage Decoder}
\begin{table}[h!]
\caption{Quantitative performance comparison to validate the efficiency of the model structures: the top two scores are highlighted in {\color[HTML]{CB0000}{\textbf{red}}} and {\bl{\textbf{blue}}}, respectively, indicating the relative advantage or disadvantage associated with each additional branch.}
\label{tab:ablation branch}
\centering
\resizebox{0.5\textwidth}{!}{
\begin{tabular}{c||c|c||cc||cc}
\toprule
                                                                       &                                                                                           &                                                                                          & \multicolumn{2}{c||}{\textbf{UIE}}                                                                                                                                                                        & \multicolumn{2}{c}{\textbf{UMDE}}                                                                                                                                                                        \\ \cline{4-7} 
\multirow{-2}{*}{}                                                     & \multirow{-2}{*}{\textbf{\begin{tabular}[c]{@{}c@{}}Infer. Time \\ (Hz)\end{tabular}}}   & \multirow{-2}{*}{\textbf{\# Params}}                                                     & \multicolumn{1}{c|}{\textbf{PSNR}}                                                                            & \textbf{SSIM}                                                                            & \multicolumn{1}{c|}{\textbf{Abs Rel}}                                                                         & \textbf{SiLog}                                                                           \\ \hline \hline
\textbf{One branch}                                                    & 122 Hz                                                                                    & 0.80 M                                                                                   & \multicolumn{1}{c|}{16.73}                                                                                    & 0.724                                                                                    & \multicolumn{1}{c|}{61.21}                                                                                    & 19.56                                                                                    \\ \hline
\textbf{Two branch}                                                    & \begin{tabular}[c]{@{}c@{}}112 Hz \\ {\color[HTML]{CB0000} \textbf{(-8.1\%)}}\end{tabular} & \begin{tabular}[c]{@{}c@{}}0.88 M\\ (+11.2\%)\end{tabular}                                & \multicolumn{1}{c|}{\begin{tabular}[c]{@{}c@{}}20.12\\ {\color[HTML]{CB0000} \textbf{(+20.2\%)}}\end{tabular}} & \begin{tabular}[c]{@{}c@{}}0.822\\ {\color[HTML]{CB0000} \textbf{(+13.5\%)}}\end{tabular} & \multicolumn{1}{c|}{\begin{tabular}[c]{@{}c@{}}47.13\\ {\color[HTML]{CB0000} \textbf{(-23.0\%)}}\end{tabular}} & \begin{tabular}[c]{@{}c@{}}14.14\\ {\color[HTML]{CB0000} \textbf{(-27.7\%)}}\end{tabular} \\ \hline
\textbf{\begin{tabular}[c]{@{}c@{}}Three branch\\ (Ours)\end{tabular}} & \begin{tabular}[c]{@{}c@{}}102 Hz \\ \bl{\textbf{(-8.9\%)}}\end{tabular} & \begin{tabular}[c]{@{}c@{}}0.96 M\\ (+9.1\%)\end{tabular}                                 & \multicolumn{1}{c|}{\begin{tabular}[c]{@{}c@{}}22.56\\ \bl{\textbf{(+12.1\%)}}\end{tabular}} & \begin{tabular}[c]{@{}c@{}}0.851\\ \bl{\textbf{(+3.5\%)}}\end{tabular}  & \multicolumn{1}{c|}{\begin{tabular}[c]{@{}c@{}}41.58\\ \bl{\textbf{(-11.8\%)}}\end{tabular}} & \begin{tabular}[c]{@{}c@{}}12.31\\ \bl{\textbf{(-12.6\%)}} \end{tabular} \\ \hline
\textbf{Four branch}                                                   & \begin{tabular}[c]{@{}c@{}}91 Hz \\ (-10.8\%)\end{tabular}                                 & \begin{tabular}[c]{@{}c@{}}1.04 M\\ \bl{\textbf{(+8.3\%)}}\end{tabular} & \multicolumn{1}{c|}{\begin{tabular}[c]{@{}c@{}}22.60\\ (+0.17\%)\end{tabular}}                                 & \begin{tabular}[c]{@{}c@{}}0.858\\ (+0.8\%)\end{tabular}                                  & \multicolumn{1}{c|}{\begin{tabular}[c]{@{}c@{}}40.11\\ (-3.5\%)\end{tabular}}                                  & \begin{tabular}[c]{@{}c@{}}12.16\\ (-1.5\%)\end{tabular}                                  \\ \hline
\textbf{Five branch}                                                   & \begin{tabular}[c]{@{}c@{}}76 Hz \\ (-16.4\%)\end{tabular}                                 & \begin{tabular}[c]{@{}c@{}}1.12 M\\ {\color[HTML]{CB0000} \textbf{(+7.7\%)}}\end{tabular} & \multicolumn{1}{c|}{\begin{tabular}[c]{@{}c@{}}22.61\\ (+0.04\%)\end{tabular}}                                 & \begin{tabular}[c]{@{}c@{}}0.861\\ (+0.3\%)\end{tabular}                                  & \multicolumn{1}{c|}{\begin{tabular}[c]{@{}c@{}}39.86\\ (-0.6\%)\end{tabular}}                                  & \begin{tabular}[c]{@{}c@{}}12.01\\ (-1.2\%)\end{tabular}                                  \\
\bottomrule
\end{tabular}
}
\end{table}
The following validation experiment demonstrates the effectiveness of our hierarchical three-stage model structure. Table \ref{tab:ablation branch} presents the quantitative results, detailing the advantages and disadvantages of using between one to five branches in terms of speed, memory, UIE, and UMDE. The percentage \texttt{values} in parentheses represent the numerical benefits and drawbacks of adding each branch. This is mathematically expressed as follows:
\begin{equation} 
    \texttt{value} (\%) = \frac{v^{n}-v^{n-1}}{v^{n}} \times 100
    \label{eq:metric for percentage}
\end{equation}
, where $n \in \left\{ 2, 3, 4, 5 \right\}$ and $v$ is the value per metric. In multi-task learning, the performance gains are most substantial when transitioning from one to two branches. Adding a third branch provides moderate additional benefits, while adding a fourth or fifth branch results in diminishing returns. Although our three-branch model does not match the two-branch model in absolute gains, it demonstrates significant advantages, outperforming other models while balancing speed and memory efficiency. Based on these findings, we propose a three-branch multi-task structure as an effective compromise between performance and resource demands.

\subsubsection{Analysis of Uncertainty Estimation Methodology}
\begin{figure}[h!]
	\centering
	\def\width{1\columnwidth}%
        {%
		\includegraphics[width=\width]{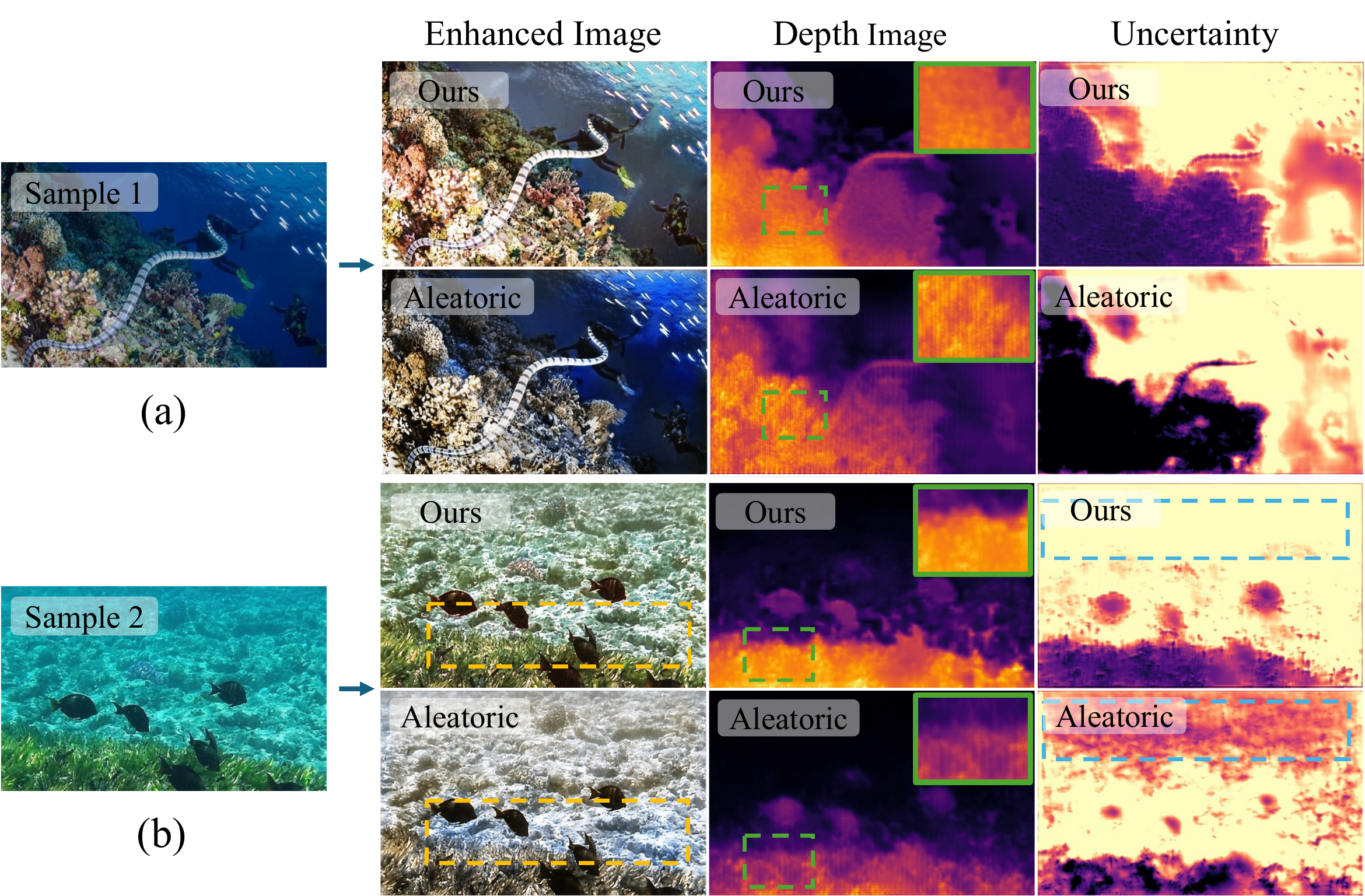}
	}
        \vspace{-3mm}
	\caption{Qualitative comparison between using aleatoric uncertainty and not using it. (a) shows our superiority over UIE and UMDE, and (b) shows our superiority over UIE, UMDE, and Uncertainty Estimation.}
	\label{fig:why_use_uncertainty2}
\vspace{-2mm}
\end{figure}

As mentioned in \ref{uncertainty extraction block}, learning aleatoric uncertainty in a small-scale multi-task model results in highly unstable optimization, as evidenced by the performance differences in the Table \ref{tab:compare aleatoric total}. In \figref{fig:why_use_uncertainty2}(a), our method and aleatoric uncertainty learning generally estimate the uncertainty well. However, when aleatoric uncertainty is simultaneously learned, the performance deteriorates on \ac{UIE} and \ac{UMDE}. In \figref{fig:why_use_uncertainty2}(b), it is shown that aleatoric uncertainty not only underperforms in UIE and UMDE but also fails to accurately estimate uncertainty.

\begin{table}[h!]
\caption{Quantitative performance comparison between Ours and Aleatoric Uncertainty: ours are \textbf{bold} in \ac{UIE} and \ac{UMDE} validation sets, respectively.}
\label{tab:compare aleatoric total}
\centering
\resizebox{0.475\textwidth}{!}{%
\begin{tabular}{c|c|c|c|c|c|c}
\toprule
\multicolumn{7}{c}{\textbf{Underwater Image Enhancement}}                                                                                                                                                                                                                                                                                                                                                                                                                                                                                                         \\ \hline
\multicolumn{1}{c||}{\multirow{2}{*}{}}                                                         & \multicolumn{2}{c||}{\textbf{Joint-ID's}}                                                                                                                    & \multicolumn{2}{c||}{\textbf{L-400}}                                                                                                                        & \multicolumn{2}{c}{\textbf{USR-248}}                                                                                                  \\ \cline{2-7} 
\multicolumn{1}{c||}{}                                                                          & \textbf{PSNR $\uparrow$}                                                     & \multicolumn{1}{c||}{\textbf{SSIM $\uparrow$}}                                                      & \textbf{PSNR $\uparrow$}                                                     & \multicolumn{1}{c||}{\textbf{SSIM $\uparrow$}}                                                     & \textbf{UIQM $\uparrow$}                                                     & \textbf{UCIQE $\uparrow$}                                                    \\ \hline \hline
\multicolumn{1}{c||}{\textbf{\begin{tabular}[c]{@{}c@{}}Aleatoric \\ Uncertainty\end{tabular}}} & 20.32                                                             & \multicolumn{1}{c||}{0.792}                                                              & 17.89                                                             & \multicolumn{1}{c||}{0.81}                                                              & 2.87                                                              & 0.612                                                             \\ \hline
\multicolumn{1}{c||}{\textbf{Ours}}                                                             & \textbf{\begin{tabular}[c]{@{}c@{}}22.56\\ (+11\%)\end{tabular}}  & \multicolumn{1}{c||}{\textbf{\begin{tabular}[c]{@{}c@{}}0.851\\ (+7.4\%)\end{tabular}}}  & \textbf{\begin{tabular}[c]{@{}c@{}}18.68\\ (+4.4\%)\end{tabular}} & \multicolumn{1}{c||}{\textbf{\begin{tabular}[c]{@{}c@{}}0.84\\ (+3.7\%)\end{tabular}}}  & \textbf{\begin{tabular}[c]{@{}c@{}}3.05\\ (+6.2\%)\end{tabular}}  & \textbf{\begin{tabular}[c]{@{}c@{}}0.638\\ (+4.2\%)\end{tabular}} \\ \hline
\multicolumn{7}{c}{\textbf{Depth Estimation}}                                                                                                                                                                                                                                                                                                                                                                                                                                                                                                                     \\ \hline \hline
\multicolumn{1}{c||}{\multirow{2}{*}{}}                                                         & \multicolumn{2}{c||}{\textbf{Joint-ID's}}                                                                                                                    & \multicolumn{2}{c||}{\textbf{Sea-thru}}                                                                                                                      & \multicolumn{2}{c}{\textbf{SQUID}}                                                                                                    \\ \cline{2-7} 
\multicolumn{1}{c||}{}                                                                          & \textbf{SiLog $\downarrow$}                                                    & \multicolumn{1}{c||}{\textbf{Sq Rel $\downarrow$}}                                                    & \textbf{SiLog $\downarrow$}                                                    & \multicolumn{1}{c||}{\textbf{Sq Rel $\downarrow$}}                                                   & \textbf{SiLog $\downarrow$}                                                    & \textbf{Sq Rel $\downarrow$}                                                   \\ \hline
\multicolumn{1}{c||}{\textbf{\begin{tabular}[c]{@{}c@{}}Aleatoric \\ Uncertainty\end{tabular}}} & 13.31                                                             & \multicolumn{1}{c||}{0.112}                                                              & 18.81                                                             & \multicolumn{1}{c||}{0.211}                                                             & 34.02                                                             & 3.68                                                              \\ \hline
\multicolumn{1}{c||}{\textbf{Ours}}                                                             & \textbf{\begin{tabular}[c]{@{}c@{}}12.35\\ (-7.2\%)\end{tabular}} & \multicolumn{1}{c||}{\textbf{\begin{tabular}[c]{@{}c@{}}0.091\\ (-18.7\%)\end{tabular}}} & \textbf{\begin{tabular}[c]{@{}c@{}}17.17\\ (-8.7\%)\end{tabular}} & \multicolumn{1}{c||}{\textbf{\begin{tabular}[c]{@{}c@{}}0.196\\ (-7.1\%)\end{tabular}}} & \textbf{\begin{tabular}[c]{@{}c@{}}33.43\\ (-1.7\%)\end{tabular}} & \textbf{\begin{tabular}[c]{@{}c@{}}2.07\\ (-43\%)\end{tabular}} \\
\bottomrule
\end{tabular}
}
\end{table}
\subsubsection{Utilization of estimated uncertainty}
\begin{figure}[b!]
	\def\width{1\columnwidth}%
        {%
		\includegraphics[width=1.025\width, right]{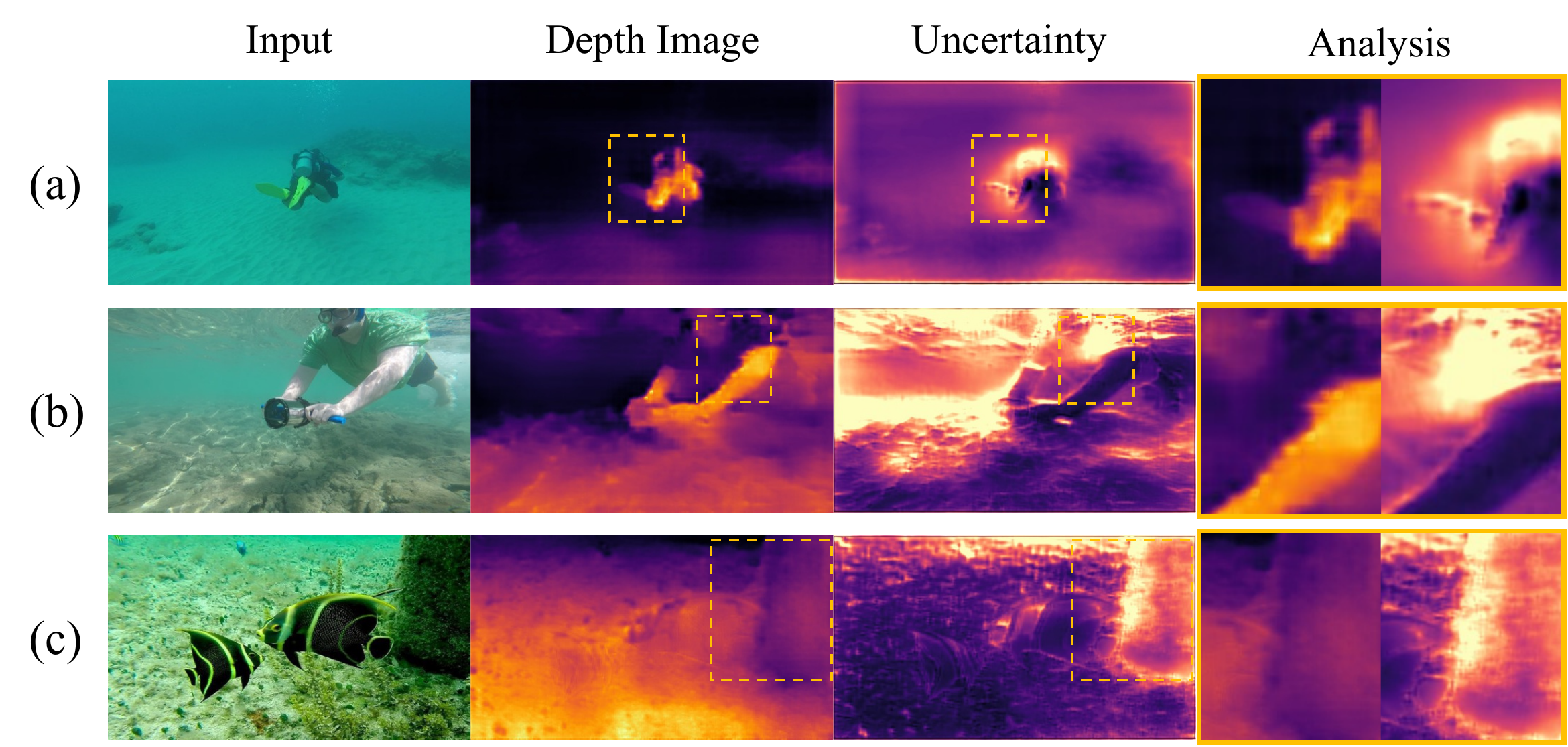}
	}
        \vspace{-3mm}
	\caption{Examples of inaccurate depth estimation for multiple depth ranges. (a) an example of failed depth estimation at a distant distance, (b) an example of failed depth estimation at a relatively close distance, and (c) an example of failed depth estimation at a close distance. Yellow boxes indicate areas where depth estimation failed, but with high uncertainty.}
	\label{fig:application uncertainty}
\end{figure}
This section demonstrates the effectiveness of incorporating uncertainty in various aspects. \figref{fig:application uncertainty} illustrates instances of depth estimation failures across different depth ranges, with subfigures (a), (b), and (c) showing relatively high uncertainty values in specific areas. These high uncertainty values assist in making informed decisions regarding the use of potentially unreliable depth estimation results. As demonstrated in Table \ref{tab:enhancement supervised}, filtering regions based on uncertainty values have been proven to enhance quantitative performance metrics.

\section{Application to robot vision} \label{Application}
\subsection{Feature Matching} \label{feature matching}
\begin{figure}[t!] %
    \centering
    \subfigure[Feature Matching Results. (top-left: Original, top-right: FUnIE-GAN \cite{islam2020fast}, bottom-left: UIEC$^2$ \cite{wang2021uiec}, bottom-right: Proposed)]{%
        \includegraphics[width=1.0\columnwidth]{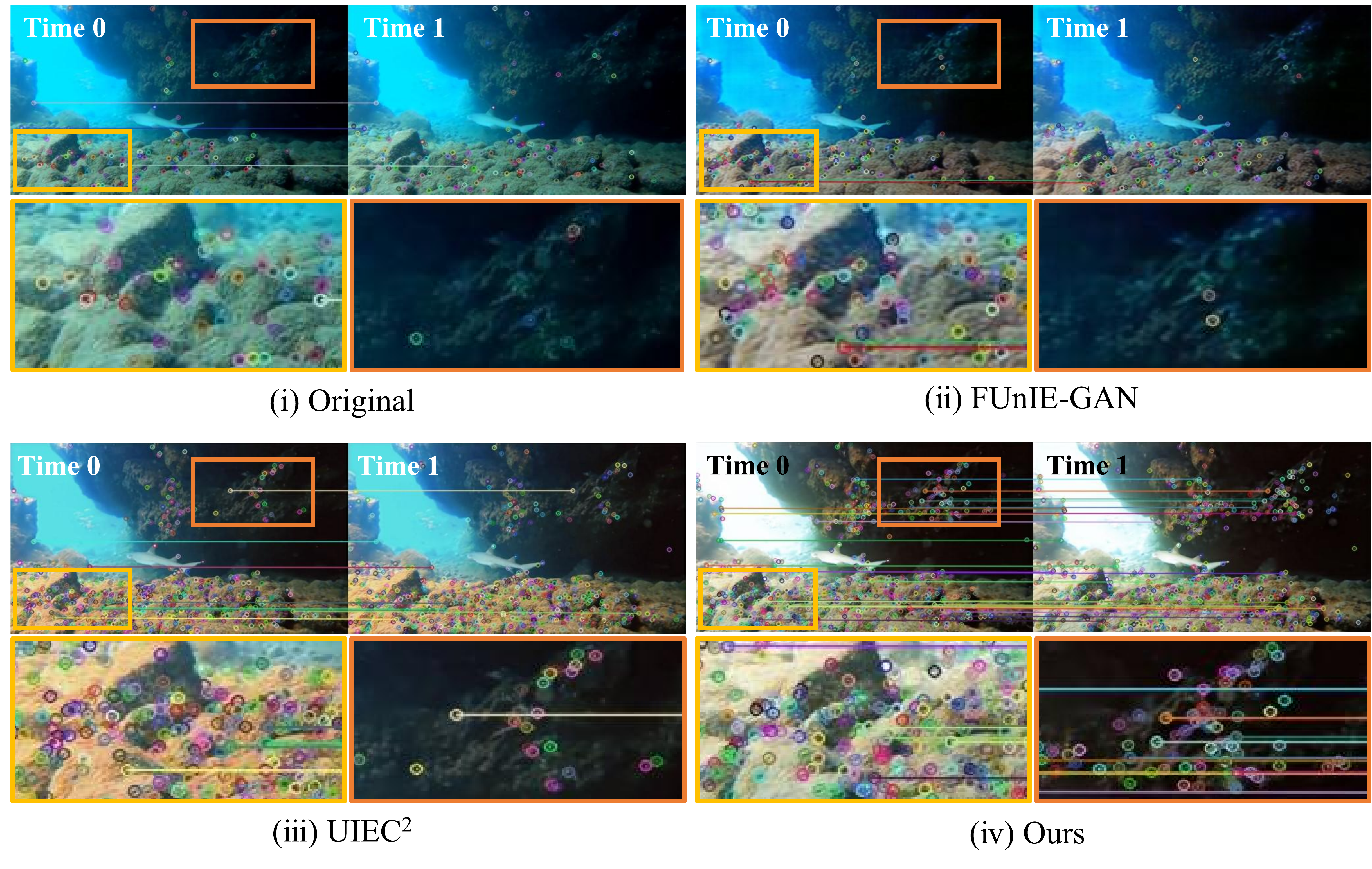}
    } \label{sub_fig_feat_match} 
    
    \subfigure[Number of Inliers for sequential images in two underwater videos]{%
        \includegraphics[width=1.0\columnwidth]{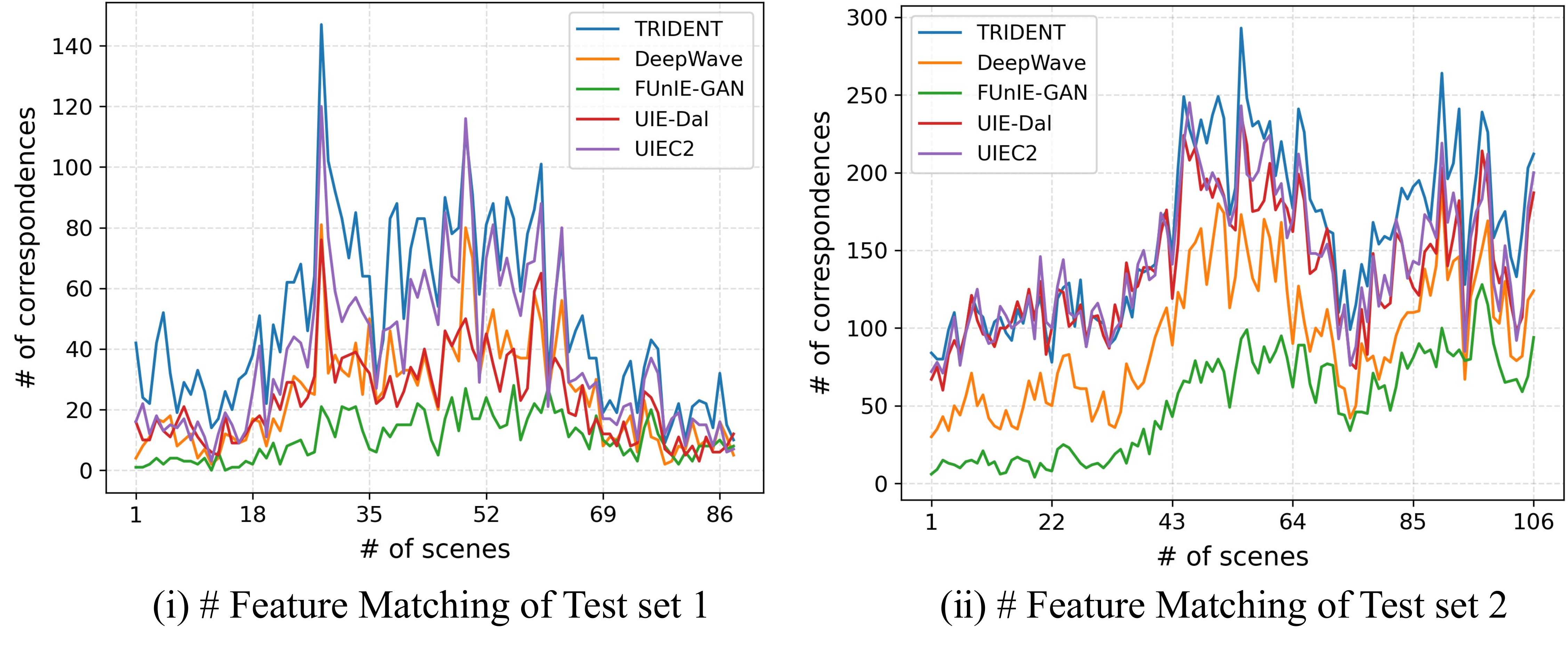}
    } \label{sub_fig_feat_graph} 
    
    \subfigure[Video samples from real underwater Test set 1, 2]{%
        \includegraphics[width=1.0\columnwidth]{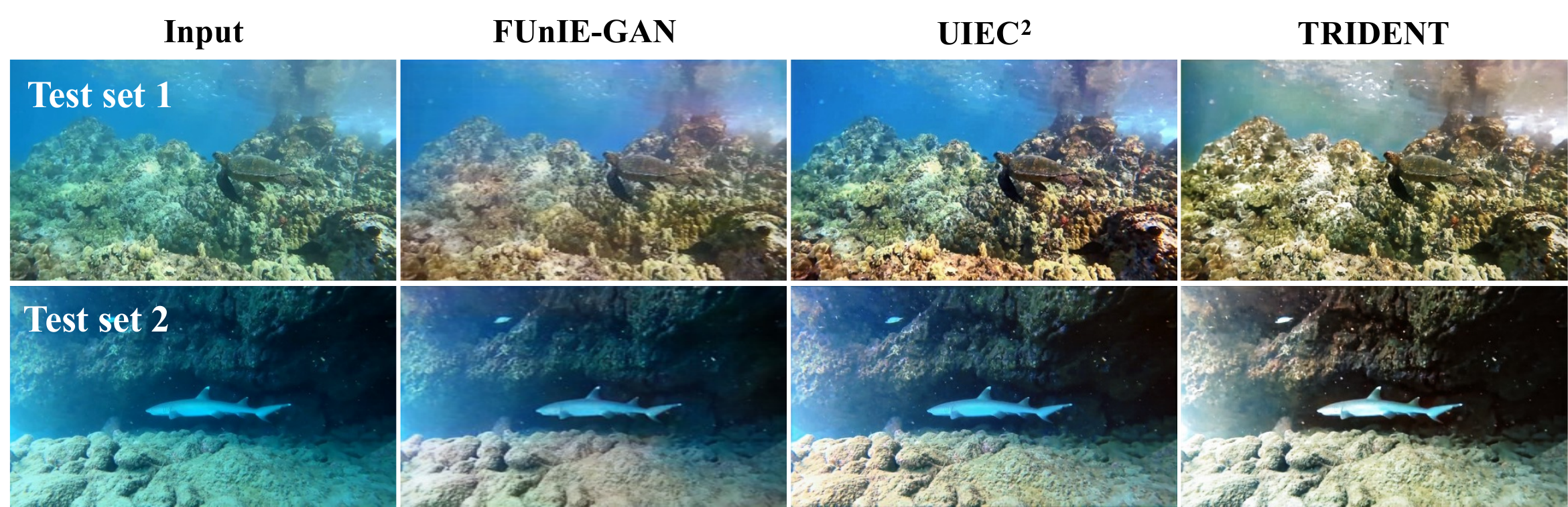}
    } \label{sub_fig_feat_video_sample} 

    \vspace{-1mm}
    \caption{Feature matching results on real underwater videos. (a) Qualitative sample of feature matching. To compare globally feature matching, we marked closed areas (yellow box) and distant areas (orange box). (b) Number of inliers and matching score on sequential images with SOTA models.}
    \label{fig:app_matching}
\vspace{-3mm}
\end{figure}

For geometric vision tasks such as SLAM and 3-D reconstruction in underwater environments, richer and more accurate feature matching is important. Our distance-independent restoration is advantageous for feature matching across the entire scene. \figref{fig:app_matching}(a) shows the qualitative results of feature matching from some models on sequential images. In the close region (yellow boxed area), both UIEC$^2$ and ours extract relatively many keypoints and obtain good data correspondences; FUnIE-GAN cannot extract many keypoints due to poor hue rectification. Our resilience is more prominent in the distant region (orange boxed area). In the orange region of each model, FUnIE-GAN and original can not extract the distant keypoints, and compared to ours, UIEC$^2$ also can not extract relatively many keypoints. The tendencies are also reflected in the number of matched inliers. \figref{fig:app_matching}(b) shows the quantitative results of feature matching on two real underwater video sets. From the graphs of $\lowercase\expandafter{\romannumeral1}$ and $\lowercase\expandafter{\romannumeral2}$, our proposed approach has the highest continued data association success rate. These results demonstrate the proposed system has strong geometry stability for 3-D robotics tasks. \figref{fig:app_matching}(c) is a sample from test sets 1 and 2.

\subsection{Salient Object Localization}
\begin{figure}[h!] %
    \vspace{-2mm}
    \centering
    \subfigure[Example of saliency detection (top: Original, bottom: Proposed)]{%
        \includegraphics[width=1.0\columnwidth]{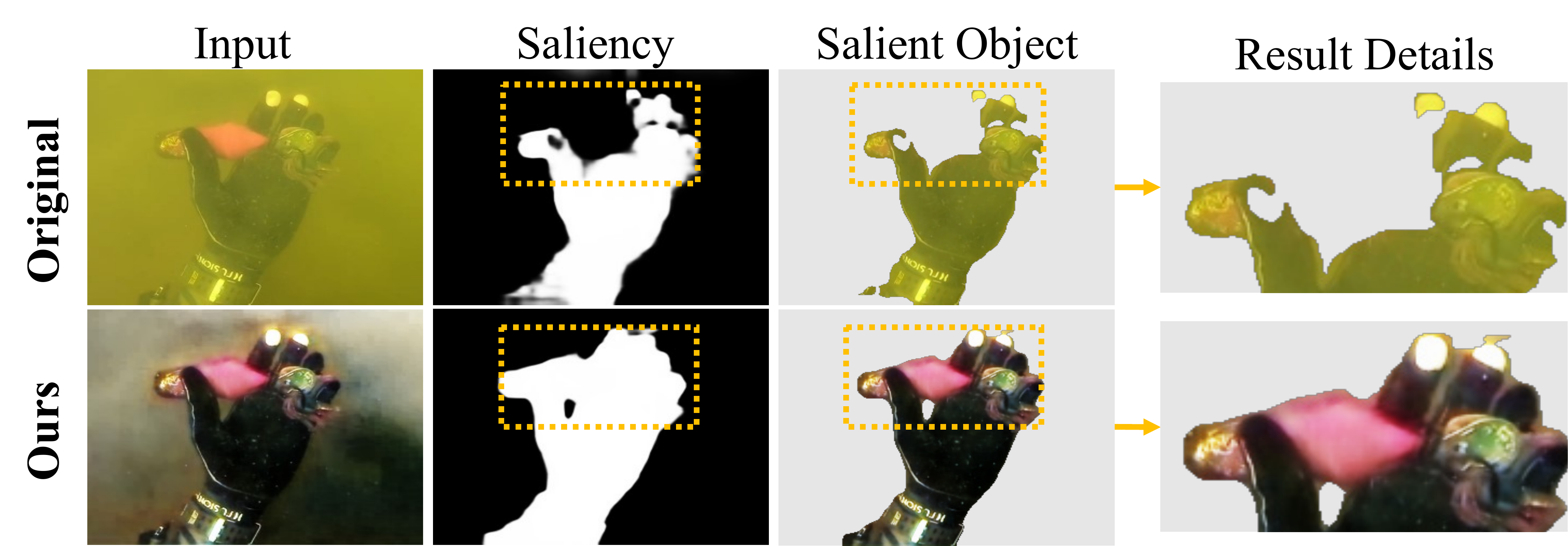}
    } \label{sub_fig_graphs} 
    \centering
    \subfigure[The process of extending to 3-D Salience Localization.]{%
        \includegraphics[width=1.0\columnwidth]{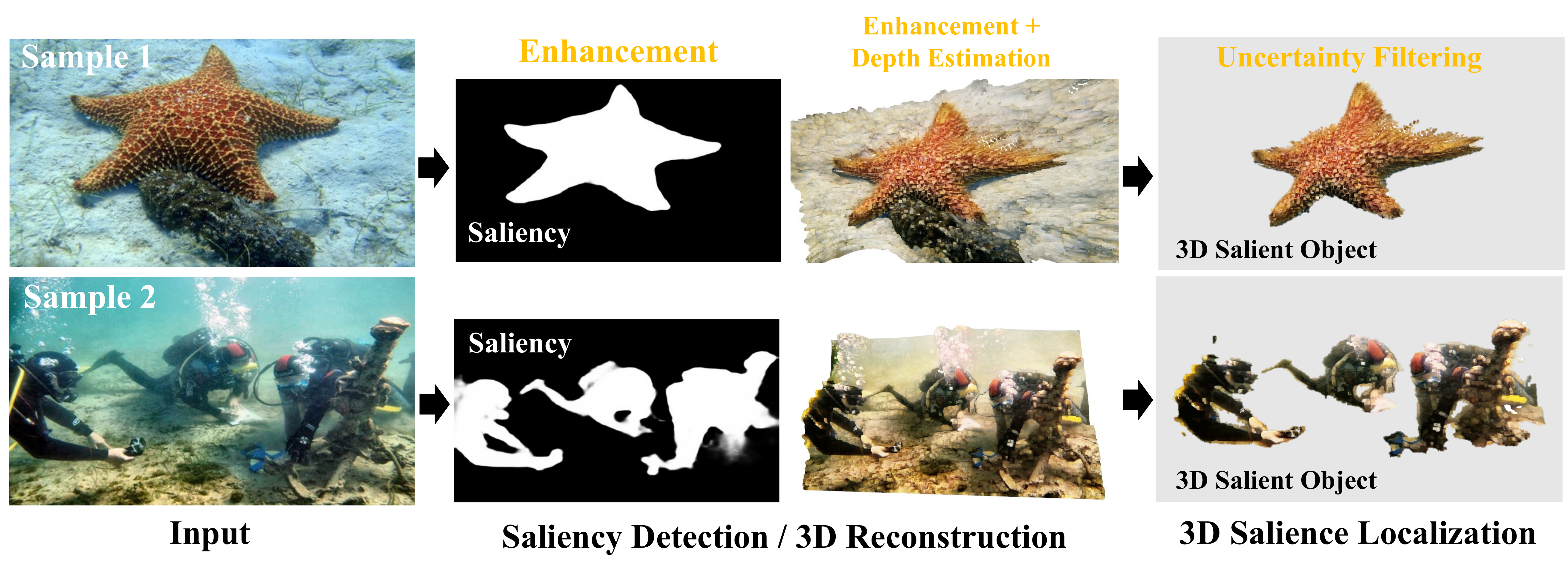}
    } \label{sub_fig_feat_match} 
    
    \caption{Examples of 3-D robot vision in an underwater scene. (a) Example of the benefit for \ac{UIE} on Saliency Detection. (b) Extending the 2-D Vision Task to 3-D Reconstruction and 3-D Salience Localization. The yellow wording is the suggested system module used in each process. We used a model in Tracer \cite{lee2022tracer} to perform saliency detection.}
    \label{fig:saliency_localization}
\end{figure}

Our proposed system combines robust image enhancement and depth estimation tasks with depth uncertainty. In this subsection, we show that the simultaneous extraction of this information enhances multiple 3-D vision tasks in both speed and accuracy; in the yellow dotted box in \figref{fig:saliency_localization}(a), the original salient object is partially erased, while our image with high color contrast retains the object's intact shape. This improvement in 2-D visual performance extends to 3-D vision tasks by utilizing depth information. For example, the extracted salient object can be extended to 3-D salience localization by recovering 3-axis coordinate information. Specifically, an example process of 3-D saliency localization is shown in \figref{fig:saliency_localization}(b); $\lowercase\expandafter{\romannumeral1})$ image enhancement is performed to make saliency detection robust, $\lowercase\expandafter{\romannumeral2})$ 3-D reconstruction of the entire scene is performed using the distance obtained through depth estimation, $\lowercase\expandafter{\romannumeral3})$ finally, the entire reconstructed scene is filtered through salient object mask and depth uncertainty. The reason for filtering through depth uncertainty is that the depth confidence of the edge of the object is low. This 3-D vision task can be greatly useful in autonomous navigation in underwater environments.

\vspace{-2mm}
\section{Conclusion}
In this paper, we propose a new training pipeline to achieve the three objectives and prove that the proposed TRIDENT has robustness in perceptual restoring, depth estimation, and uncertainty at the \ac{SOTA}; we showed it outperforms in quantitative and qualitative \ac{UIE}, \ac{UMDE} evaluation. We also confirmed that the overall depth estimation and image enhancement accuracy can be increased through uncertainty filtering. we demonstrated it is light-weight and fast compared to other \ac{UMDE} and \ac{UIE} models, despite performing two or three tasks. In addition, the depth-wise image restoration approach has proven to be advantageous in obtaining rich correspondences in feature matching. Finally, we demonstrated that our mechanism is not limited to extending from 2-D to 3-D robot vision systems. However, this study has a limitation: the depth scale is unknown. It can create a scale problem for large-scale 3-D reconstruction or SLAM. In the future, we would like to solve this problem through depth completion or sonar sensor fusion.

 \section*{Acknowledgment}
 This work was supported by the National Research Foundation of Korea (NRF) grant (No. RS-2023-00302589) and the Institute of Information \& communications Technology Planning \& Evaluation (IITP) grant funded by the Korean government(MSIT) (No.2022-0-00448/RS-2022-II220448). Also, this work is supported by Korea Institute of Marine Science and Technology Promotion (KIMST), funded by the Ministry of Oceans and Fisheries (20210562).

\small 
\bibliographystyle{packages/IEEEtranN} %
\bibliography{packages/string-short, packages/references}

\begin{IEEEbiography}[{\includegraphics[width=1in,height=1.25in,clip,keepaspectratio]{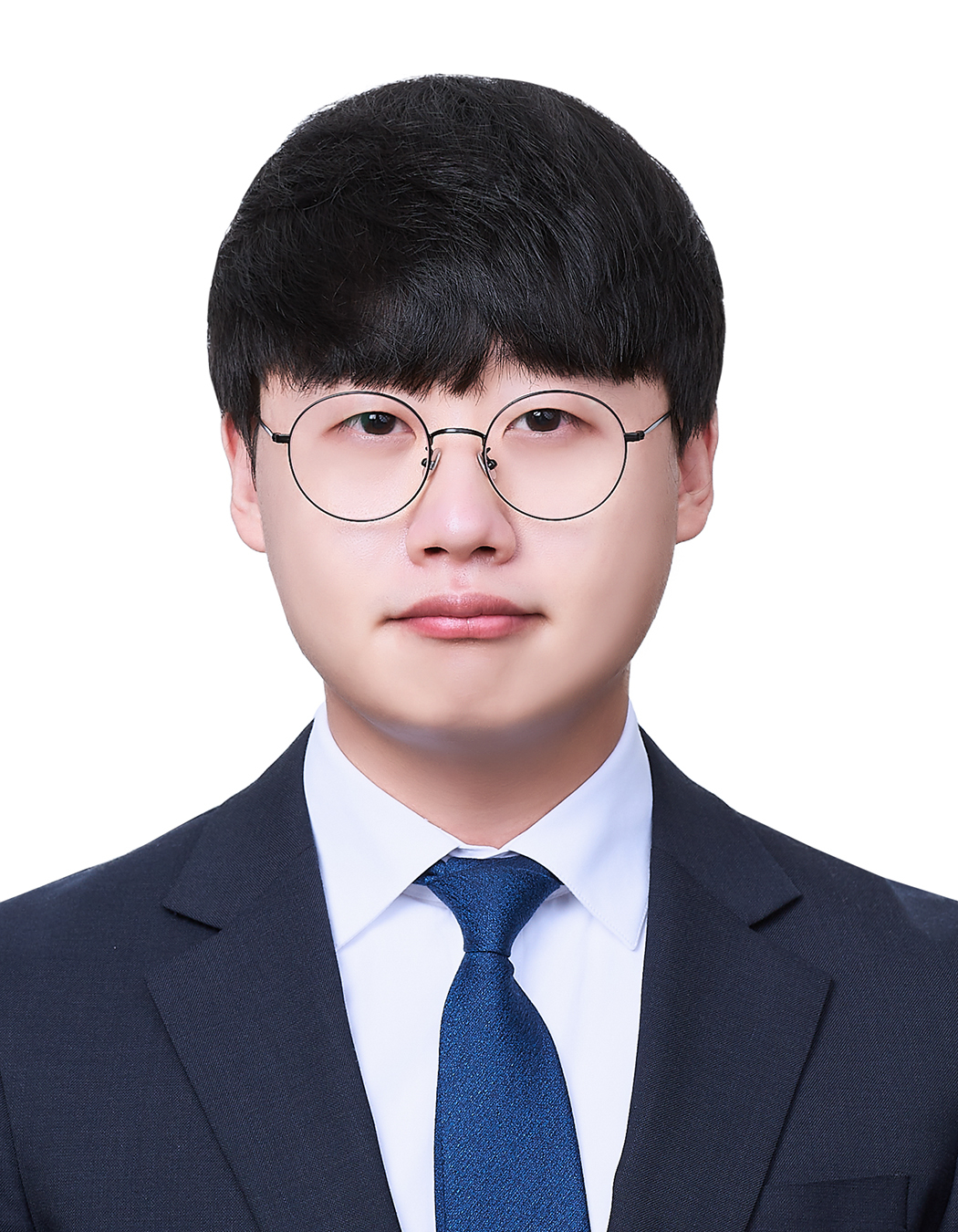}}]{Geonmo Yang} (Graduate Student Member, IEEE) received the B.S. degree in the electronic engineering from Tech University of Korea (TUKOREA), Siheung, South Korea, in 2022. Currently, he is Ph.D. student in the electrical and computer engineering from Inha University, Incheon, Korea. His research interests include visual-lidar simultaneous localization and mapping (SLAM), computer vision for robust sensing and graphics, and robotics for long-term autonomy.
\end{IEEEbiography}

\begin{IEEEbiography}[{\includegraphics[width=1in,height=1.25in,clip,keepaspectratio]{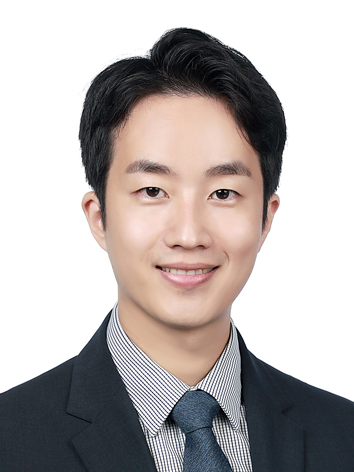}}]{Younggun Cho} (Member, IEEE) received the B.S. degree in electrical engineering from Inha University, Incheon, Korea, in 2013, and the M.S. degree in electrical engineering in 2015 and Ph.D. degree in civil and environmental engineering with dual degree in robotics program (2020) from the Korea Advanced Institute of Science \& Technology (KAIST), Daejeon, Korea. He was an Assistant Professor with the department of Robotics Engineering at Yeungnam University. He is currently an Assistant Professor with the department of Electrical Engineering at Inha University, Incheon, South Korea. His current research interests include robust sensing, long-term autonomy, and visual simultaneous localization and mapping.
\end{IEEEbiography}

\end{document}